\documentclass[a4paper,10pt]{amsart}

\usepackage{cite}
\usepackage{amsthm,amssymb}
\usepackage{mathtools,thmtools,thm-restate}
\usepackage{bm,esint}
\usepackage{color}
\usepackage{float,graphicx,subcaption}
\usepackage{cancel}
\usepackage[shortlabels]{enumitem}
\usepackage{mathrsfs}  
\usepackage{bbm}
\usepackage{euscript}
\usepackage{hyperref}
\usepackage{cleveref}
\usepackage{upgreek}
\usepackage{tikz-cd}
\usetikzlibrary{decorations.pathmorphing}

\usepackage{todonotes}

\usepackage{algorithm}
\usepackage{algpseudocode}

\usepackage{graphicx} 
\usepackage{booktabs} 
\usepackage{verbatim} 
\usepackage{multimedia}
\usepackage{subcaption}
\usepackage{adjustbox}

\declaretheoremstyle[
  headfont=\normalfont\bfseries\itshape,
  numbered=yes,
  bodyfont=\normalfont,
  spaceabove=1em plus 0.75em minus 0.25em,
  spacebelow=.5em,
  qed={\small$\Diamond$},
]{deflt}

\declaretheorem[style=deflt,numberwithin=section]{theorem}

\declaretheorem[style=deflt,sibling=theorem]{proposition}
\declaretheorem[style=deflt,sibling=theorem]{corollary}

\declaretheorem[style=deflt,sibling=theorem]{remark}

\numberwithin{equation}{section}

\newcommand{\R}{\mathbb{R}}
\newcommand{\C}{\mathbb{C}}
\newcommand{\E}{\mathbb{E}}
\newcommand{\N}{\mathbb{N}}

\newcommand{\Z}{\mathbb{Z}}

\newcommand{\cE}{\mathcal{E}}
\newcommand{\cF}{\mathcal{F}}

\newcommand{\cK}{\mathcal{K}}
\newcommand{\cL}{\mathcal{L}}

\newcommand{\cP}{\mathcal{P}}
\newcommand{\cQ}{\mathcal{Q}}
\newcommand{\cR}{\mathcal{R}}

\newcommand{\cU}{\mathcal{U}}
\newcommand{\cV}{\mathcal{V}}

\newcommand{\cX}{\mathcal{X}}
\newcommand{\cY}{\mathcal{Y}}

\newcommand{\size}{\mathrm{size}}

\renewcommand{\hat}{\widehat}
\renewcommand{\bar}{\overline}

\newcommand{\set}[2]{{\left\{ #1 \,\middle|\, #2 \right\}}}
\newcommand{\slot}{{\,\cdot\,}}
\newcommand{\Lip}{\mathrm{Lip}}

\newcommand{\Err}{\mathscr{E}}

\usepackage{dsfont}
\usepackage{tikz,pgfplots}
\usetikzlibrary{patterns}

\newcommand{\Psid}{\Psi^\dagger}

\newcommand{\dOmega}{\mathfrak{D}}
\newcommand{\dD}{\mathfrak{D}}

\newcommand{\muN}{\mu_{{\tiny N}}}

\newcommand{\dx}{{d_\cU}}
\newcommand{\dy}{{d_\cV}}
\newcommand{\dc}{{d_c}}
\newcommand{\kmax}{k_{\mathrm{max}}}

\newcommand{\ttheta}{\gamma}

\definecolor{mypink1}{rgb}{0.75, 0.0, 0.0}

\title[Operator Learning]{Operator Learning: Algorithms and Analysis}
\author{Nikola B. Kovachki \and Samuel Lanthaler \and Andrew M. Stuart}
\date{ \today }

\makeatletter
\renewcommand{\paragraph}{%
  \@startsection{paragraph}{4}%
  {\z@}{.8ex \@plus 1ex \@minus .2ex}{-1em}%
  {\normalfont\normalsize\bfseries}%
}
\makeatother

\usepackage{color}
\definecolor{darkred}{rgb}{.6,0,0}
\definecolor{darkblue}{rgb}{0,0,.7}
\definecolor{darkgreen}{rgb}{0,.7,0}
\definecolor{darkbrown}{rgb}{0.8,0.4,0.4}

\newtheoremstyle{named}{}{}{\itshape}{}{\bfseries}{.}{.5em}{\thmnote{#3}}
\theoremstyle{named}

\begin{document}

\begin{abstract}
Operator learning refers to the application of ideas from machine learning to approximate 
(typically nonlinear) operators mapping between Banach spaces of functions. Such operators often arise from physical models expressed in terms of partial differential equations (PDEs). In this context, such
approximate operators hold great potential as efficient surrogate models to complement traditional numerical methods in many-query tasks. Being data-driven, they also enable model discovery when a mathematical description in terms of a PDE is not available. This review focuses primarily on neural operators,
built on the success of deep neural networks in the approximation of functions defined on finite dimensional
Euclidean spaces. Empirically, neural operators have shown success in a variety of applications, but our theoretical understanding remains incomplete. This review article summarizes recent progress and the current state of our theoretical understanding of neural operators, focusing on an approximation theoretic point of view.
\end{abstract}

\maketitle

\section{Introduction}
\label{sec:I}

This paper overviews algorithms and analysis related to the subject of operator learning:
finding approximations of maps between Banach spaces, from data. Our focus is primarily on
neural operators, which leverage the success of neural networks in finite dimensions; but
we also cover related literature in the work specific to learning linear operators,
and the use of Gaussian processes and random features. In subsection \ref{ssec:H} we
discuss the motivation for our specific perspective on operator learning. Subsection
\ref{ssec:L} contains a literature review. Subsection \ref{ssec:Ov} overviews the
remainder of the paper.

\subsection{High Dimensional Vectors Versus Functions}
\label{ssec:H}

Many tasks in machine learning require operations on high dimensional
tensors\footnote{``Tensor'' here may be a vector, matrix or object with more than two indices.} arising, for example, from pixellation of images or from discretization 
of a real-valued mapping defined over a subset of $\R^d.$ The main idea underlying 
the work that we overview in this
paper is that it can be beneficial, when designing and analyzing algorithms in this context, 
to view these high dimensional vectors as functions $u: \dOmega \to \R^c$ defined on a domain $\dOmega \subset \R^d$. For example  $(c,d)=(3,2)$ for RGB images and $(c,d)=(1,3)$ 
for a scalar field such as temperature in a room. Pixellation, or discretization, of $\dOmega$ will lead to a tensor with size scaling like $N$, the number of
pixels or discretization points; $N$ will be large and hence the tensor
will be of high dimension.  Working with data-driven algorithms designed to act on function $u$, rather than the high dimensional tensor, captures intrinsic properties of the problem,
and not details related to specific pixellation or discretization; as a consequence
models learned from data can be transfered from
one pixellation or discretization level to another.

\begin{figure}[h!]
    \centering
    \begin{subfigure}{.45\textwidth}
         \includegraphics[width=.94\textwidth]{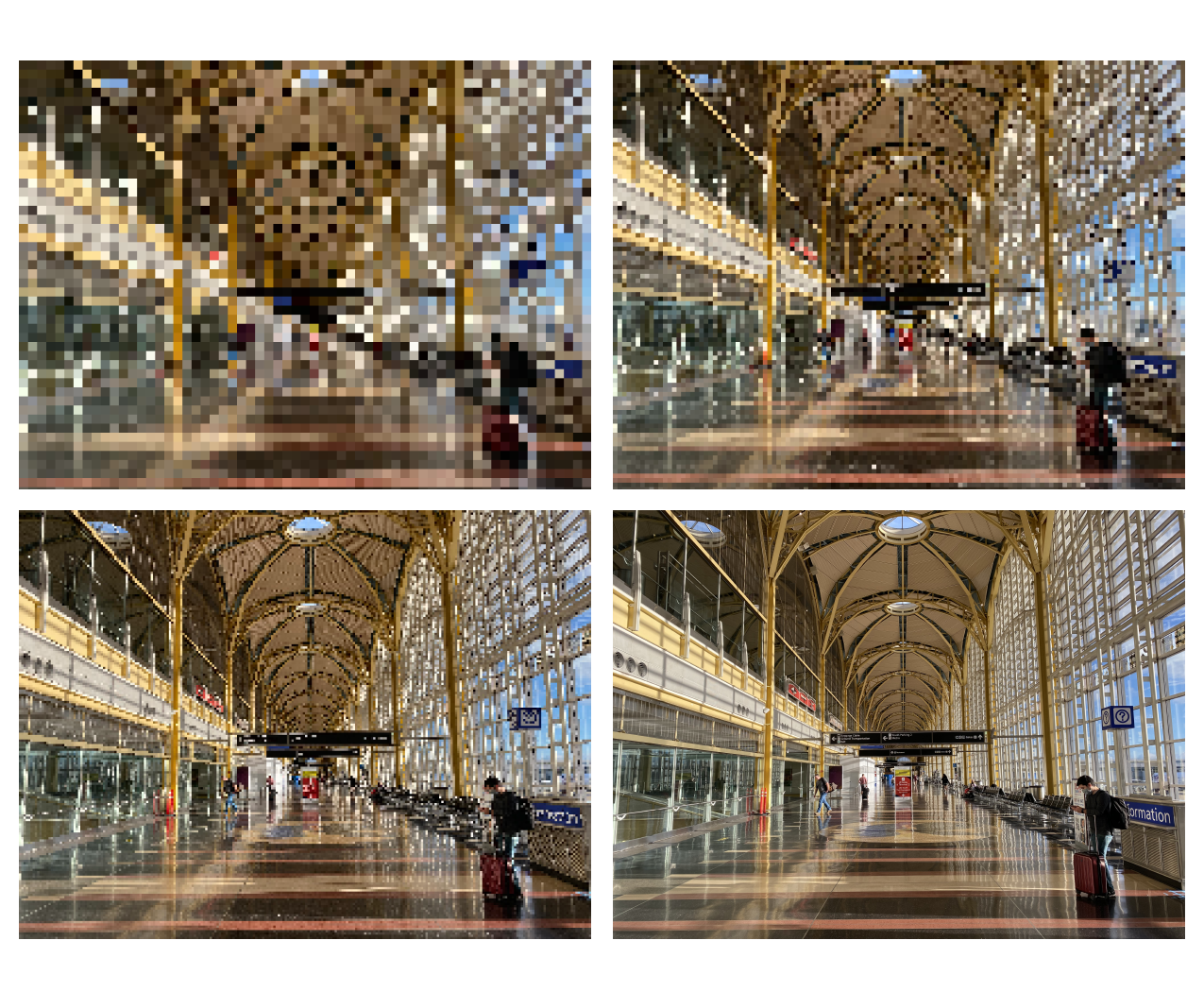}
        \caption{Same images at different resolutions}
        \label{f:1a}
    \end{subfigure}
    \hspace{1em}
    \begin{subfigure}{.45\textwidth}
       \includegraphics[width=.75\textwidth]{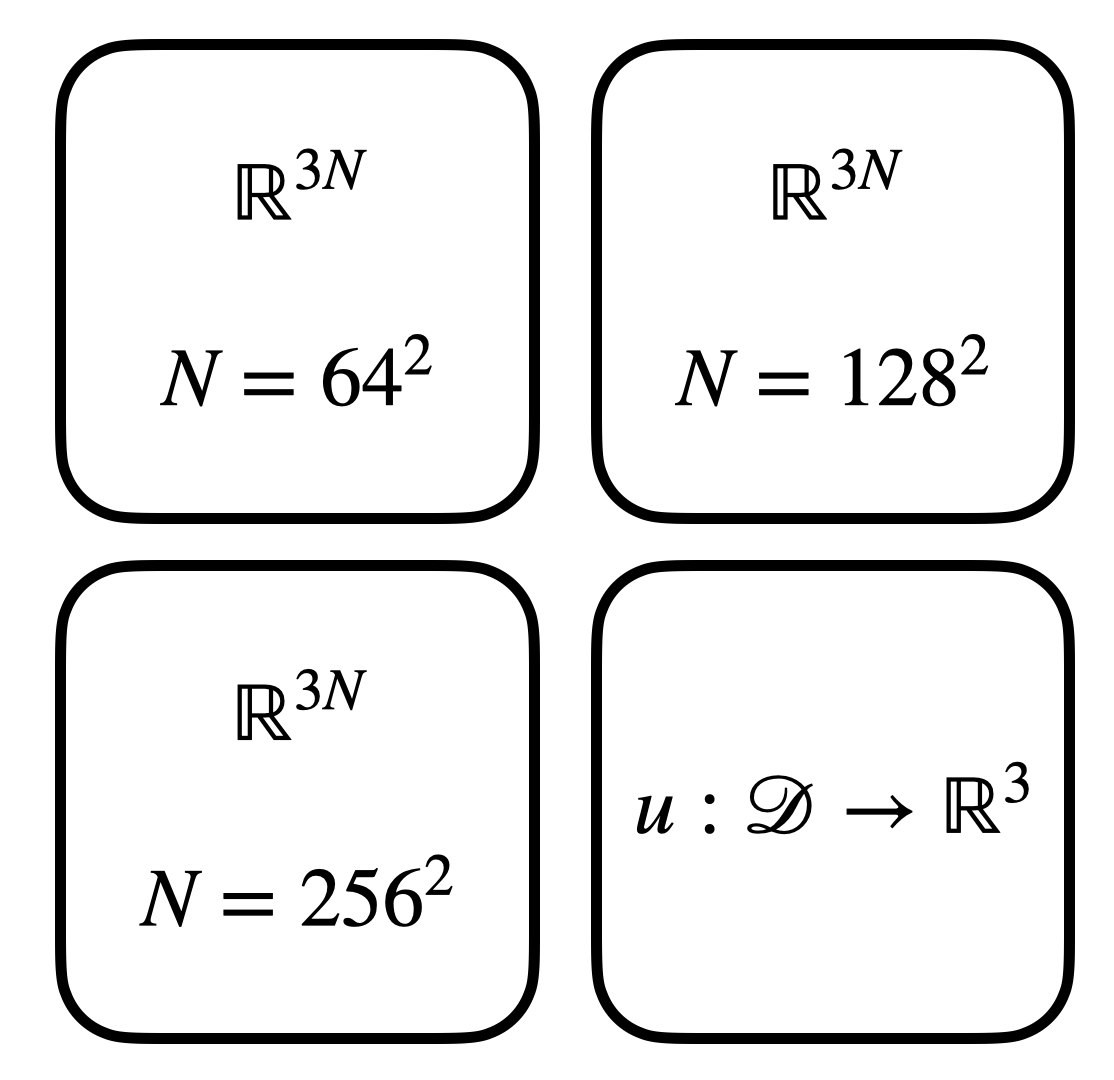}
        \caption{Different resolutions as vectors and (bottom right) as a function}
        \label{f:1b}
    \end{subfigure}
\end{figure}

\begin{figure}[h!]
    \centering
    \begin{subfigure}{.45\textwidth}
        \centering
        \includegraphics[width=.9\textwidth]{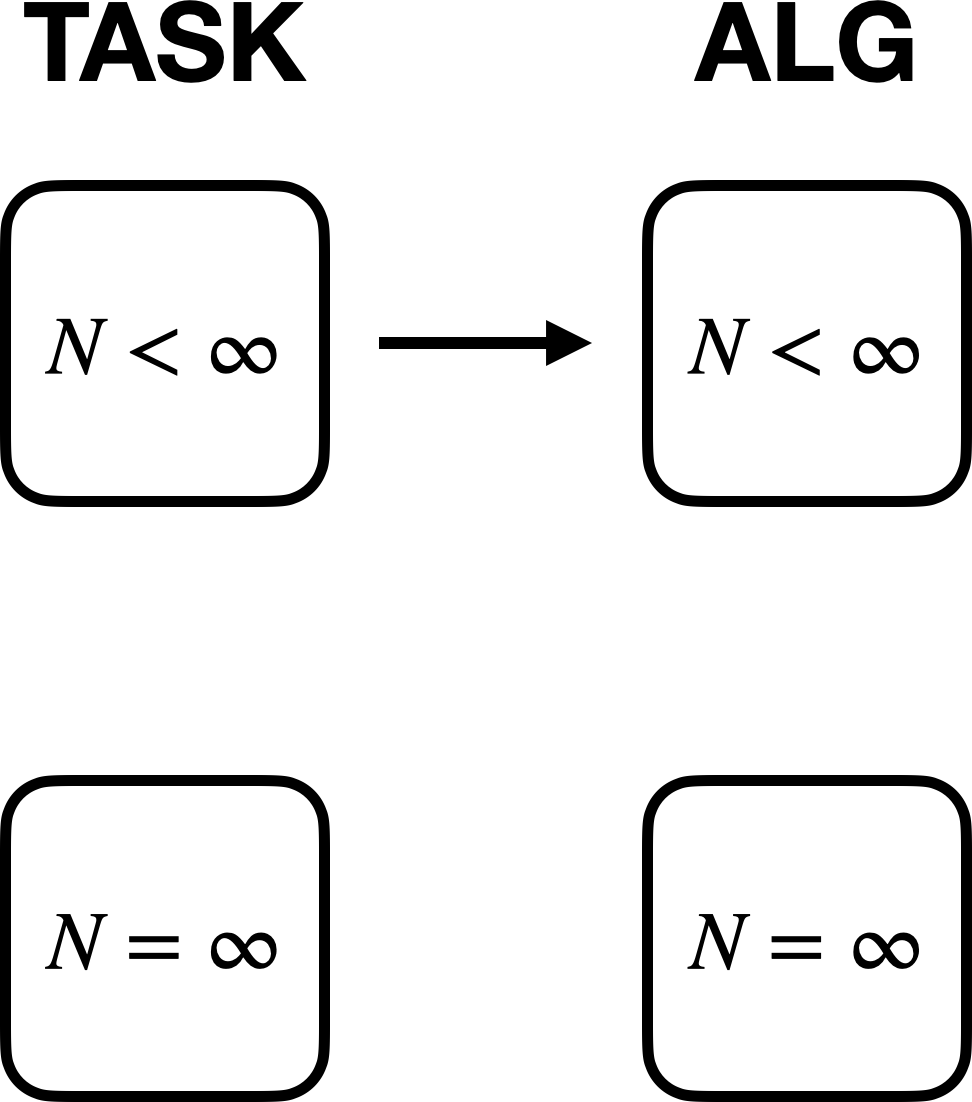}
        \caption{Directly design algorithm at fixed resolution $N$}
        \label{f:2a}
    \end{subfigure}
    \hspace{1em}
    \begin{subfigure}{.45\textwidth}
        \centering
        \includegraphics[width=.9\textwidth]{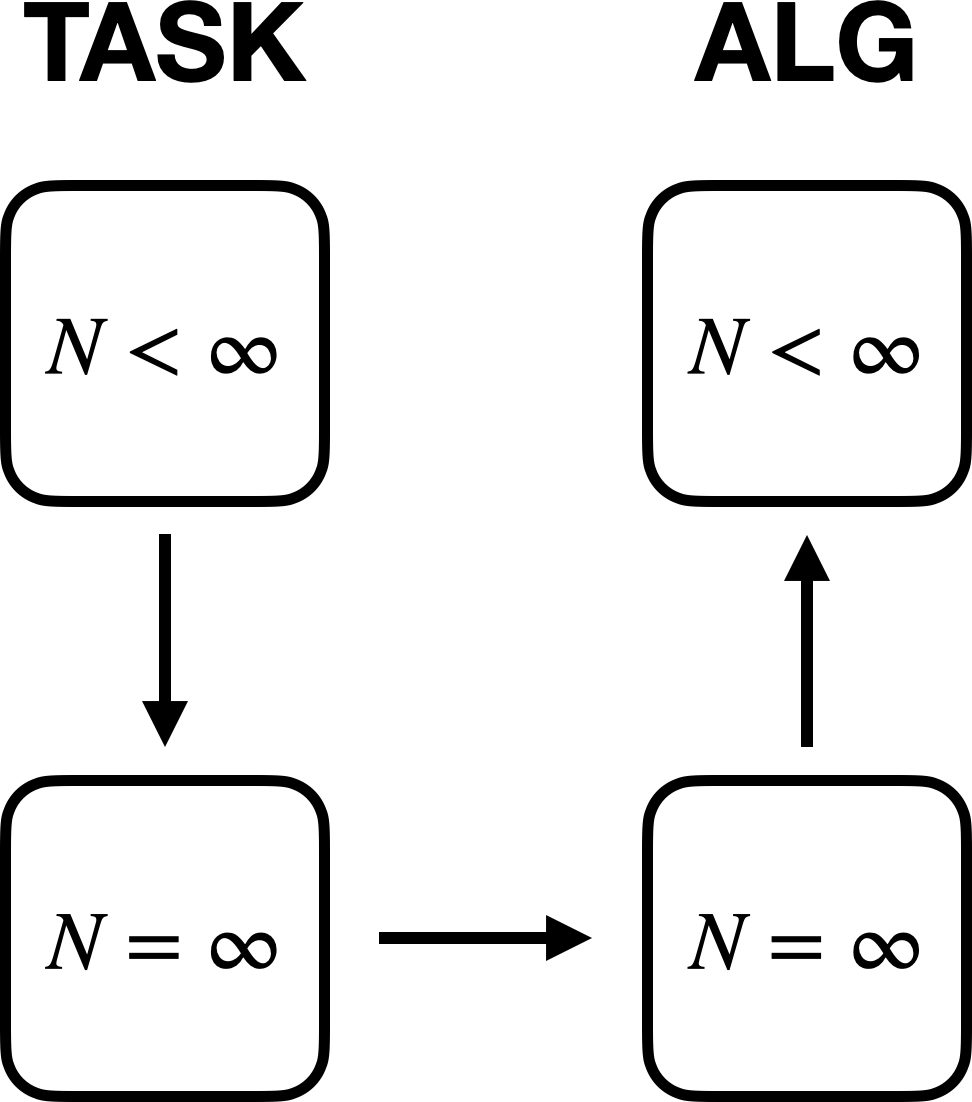}
        \caption{Design algorithm at limit of infinite resolution}
        \label{f:2b}
    \end{subfigure}
\end{figure}

Consider the image shown in Figure \ref{f:1a}, at four levels of resolution. 
As an RGB image it may be viewed as a vector in $\R^{3N}$ where $N$ is the number of pixels.
However by the time we reach the highest resolution (bottom right) it is more 
instructive to  
view it as a function mapping $\dOmega:=[0,1]^2 \subset \R^2$ into $\R^3.$
This idea is summarized in Figure \ref{f:1b}.
Even if the original machine learning task presents as acting on high dimensional tensor of dimension proportional to $N$, it is worth considering whether it may be formulated in the continuum limit $N \to \infty$,
conceiving of algorithms in this setting, and only then approximating to finite dimension again
to obtain practical algorithms. These ideas are illustrated in Figures \ref{f:2a} and \ref{f:2b}.

\subsection{Literature Review}
\label{ssec:L}

We give a brief overview of the literature in this field; greater depth, and more citations, are given in subsequent sections.

\paragraph{Algorithms on Function Space}
The idea of conceiving algorithms in the continuum, and only then discretizing,
is prevalent in numerous areas of computational science and engineering.
For example in the field of PDE constrained optimization the relative merits
of optimize-then-discretize, in comparison with discretize-then-optimize,
are frequently highlighted \cite{hinze2008optimization}.
In the field of Bayesian inverse problems \cite{kaipio2006statistical} formulation 
on Banach space \cite{stuart2010inverse} leads to new perspectives on algorithms, for
example in MAP estimation \cite{klebanov2023aperiodic}. And sampling probability
measures via MCMC can be formulated on Banach spaces \cite{cotter2013mcmc}, 
leading to provably dimension independent convergence rates \cite{hairer2014spectral}.

\paragraph{Supervised Learning on Function Space}
Supervised learning \cite{goodfellow2016deep} rose to prominence in the context of the use
of deep neural network (DNN) methods for classifying digits and then images. 
In such contexts the task is formulated as learning a mapping from a Euclidean space
(pixellated image) to a set of finite cardinality. Such methods are readily generalized to
regression in which the output space is also a Euclidean space. However, applications in
science and engineering, such as surrogate modeling \cite{sacks} and scientific discovery
\cite{raghu2020survey}, often suggest supervised learning tasks in which input and/or output
spaces are infinite dimensional; in particular they comprise spaces of functions defined
over subsets of Euclidean space. We refer to the resulting methods, conceived to solve supervised learning tasks where the inputs and outputs are functions, as neural operators. Whilst there is earlier work on regression in function space \cite{ramsay1991some}, perhaps the earliest paper to conceive of neural network-based supervised learning between spaces of functions is \cite{chen1995universal}. This work was generalized in the seminal
DeepONet paper \cite{lu2021learning}. Concurrently with the development of DeepONet other
methods were being developed, including methods based on model reduction \cite{bhattacharya2021model} (PCA-Net)
and on random features \cite{nelsen2021random}. 
The random feature approach in \cite{nelsen2021random}
included the use of manipulations in the Fourier domain, to learn the solution operator for
viscous Burgers' equation, whose properties are well-understood in Fourier space.  We also mention a related Fourier-based approach in \cite{patel2018nonlinear,patel2021physics}. The idea  of using the Fourier transform was exploited more systematically through development of  the Fourier Neural Operator (FNO) \cite{li2021fourier}. The framework introduced in this paper has subsequently been generalized to work with sets of functions other than Fourier, such as wavelets \cite{tripura2023wavelet}, spherical harmonics \cite{bonev2023spherical} and more general  sets of functions \cite{benitez2023outofdistributional,lanthaler2023nonlocal}. The FNO architecture is related to convolutional neural networks, which have also been explored for operator learning, see e.g. \cite{raonic2023,raonic2023convolutional,franco2023approximation,lippe2024pderefiner}, and \cite{rahman2023uno,gupta2022towards} for relevant work. We mention also similar developments in computer graphics where, in \cite{ovsjanikov2012functional}, a method is proposed based on projections onto the eigenfunctions of the Laplace-Beltrami operator and is subsequently extended in \cite{yi2017syncspeccnn, litany2017deep, sharp2022diffusionnet}. At a more foundational level, recent work \cite{bartolucci2023neural} develops a theoretical framework to study neural operators, aiming to pinpoint theoretical distinctions between these infinite-dimensional architectures from conventional finite-dimensional approaches, based on a frame-theoretic notion of representation equivalence.

\paragraph{Approximation Theory} 
The starting point for approximation theory is universal approximation. Such theory
is overviewed in the finite dimensional setting in \cite{pinkus1999approximation}.
It is developed systematically for neural operators in \cite{kovachki2022machine},
work that also appeared in \cite{kovachki2023neural}. However, the first paper 
to study universal approximation, in the context of mappings between spaces of scalar-valued
functions, is Chen and Chen \cite{chen1995universal}. This was followed by work extending their analysis to DeepONet \cite{lanthaler2022error}, 
analysis for the FNO \cite{kovachki2021universal} and analysis for PCA-Net in \cite{bhattacharya2021model}, and a number of more recent contributions, e.g. \cite{zhang2023belnet,HUA202321,jin2022mionet,castro2023kolmogorov,castro2022calder,huang2024operator}.

The paper \cite{lu2021learning} first introduced DeepONets and studied their practical application on a number of prototypical problems involving differential equations.
The empirical paper \cite{de2022cost} studies various neural operators from the perspective of
the cost-accuracy trade-off, studying how many parameters, or how much data, is needed
to achieve a given error. Such complexity issues are studied theoretically
for DeepONet, in the context of learning the solution operator for the
incompressible Navier-Stokes equation and several other PDEs, in \cite{lanthaler2022error}, with analogous analysis for PCA-Net in \cite{lanthaler2023operator}. In \cite{marcati2023exponential}
the coefficient to solution map is studied for divergence form elliptic
PDEs, and analyticity of the coefficient and the solution
is exploited to study complexity of the resulting neural operators.
The paper \cite{herrmann2022neural} studies complexity for the same problem,
but exploits operator holomorphy. In \cite{lanthaler2023curse} complexity
is studied for  Hamilton-Jacobi equations, using approximation of the underlying
characteristic flow. The work \cite{furuya2024globally} discusses conditions under which neural operator layers are injective and surjective. The sample complexity of operator learning with DeepONet and related architectures is discussed in \cite{liu2024deep}. Out-of-distribution bounds are discussed in \cite{benitez2023outofdistributional}.

The paper \cite{de2023convergence} studies the learning of linear operators from data.
This subject is developed for elliptic and parabolic equations, and in particular
for the learning of Greens functions, in \cite{gin2020deepgreen,boulle2022data,boulle2022learning,boulle2023learning,wang2023operator,stepaniants2023learning} and, for spectral properties of 
the Koopman operator, the solution
operator for advection equations, in \cite{kostic2022learning,colbrook2024rigorous}.

\subsection{Overview of Paper}
\label{ssec:Ov}
In section \ref{sec:O} we introduce operator learning as a supervised learning problem on
Banach space; we formulate testing and training in this context, and provide an example from
porous medium flow. Section \ref{sec:Sp} is devoted to definitions of the supervised learning
architectures that we focus on in this paper: PCA-Net, DeepONet, the Fourier Neural Operator 
(FNO) and random features methods. Section \ref{sec:AT} describes various aspects of universal approximation
theories in the context of operator learning. In section \ref{sec:Q} we study complexity of 
these approximation, including discussion of linear operator learning; specifically we
study questions such as how many parameters, or how much data, is required to achieve an
operator approximation with a specified level of accuracy; and what properties of the operator
can be exploited to reduce complexity? We summarize and conclude in section \ref{sec:C}.


\section{Operator Learning}
\label{sec:O}

In subsection \ref{ssec:S} we define supervised learning, followed in subsection \ref{ssec:S2} by discussion
of the topic in the specific case of operator learning. Subsection \ref{ssec:T} is devoted to explaining how the approximate operator is found from data (training) and how it is evaluated (testing). Subsection \ref{ssec:FLS} describes how latent structure can be built into operator approximation, and learned
from data. Subsection \ref{ssec:ExD} contains an example from parametric partial differential equations (PDEs)
describing flow in a porous medium.

\subsection{Supervised Learning}
\label{ssec:S}
The objective of supervised learning is to determine an underlying mapping $\Psid : \cU \rightarrow \cV$ from samples\footnote{Note that from now on $N$ denotes the data volume (and not the size of a finite dimensional problem as in subsection \ref{ssec:H}).}
\begin{equation}
    \label{eq:data}
\{u_n,\Psid(u_n)\}_{n=1}^N,\quad u_n \sim \mu.
\end{equation}
Here the probability measure $\mu$ is supported on $\cU$.
Often supervised learning is formulated by use of the data model
\begin{equation}
    \label{eq:data2}
\{u_n,v_n\}_{n=1}^N,\quad (u_n,v_n) \sim \pi,
\end{equation}
where the probability measure $\pi$ is supported on $\cU \times \cV.$ The
data model \eqref{eq:data} is a special case which is
sufficient for the exposition in this article.

In the original applications of supervised learning
$\cU=\R^{d_{x}}$ and $\cV=\R^{d_{y}}$ (regression) or
$\cV=\{1,\dotsm K\}$ (classification). We now go beyond this
setting.

\subsection{Supervised Learning of Operators}
\label{ssec:S2}
In many applications arising in science and engineering it is
desirable to consider a generalization of the finite-dimensional setting to separable 
Banach spaces $\cU,\cV$ of vector-valued functions:
\begin{align*}
    \cU &= \{u: \dOmega_x \rightarrow \R^{d_i}\}, \quad  \dD_x \subseteq \R^{d_x} \\
    \cV &= \{v: \dOmega_y \rightarrow \R^{d_o}\}, \quad  \dD_y \subseteq \R^{d_y}.
\end{align*}
Given data \eqref{eq:data} we seek to determine an approximation to $\Psid: \cU \to \cV$
from within a family of parameterized functions
$$
\Psi: \cU \times \Theta \mapsto \cV.
$$
Here $\Theta \subseteq \R^p$ denotes the parameter space from which we seek the optimal choice of parameter, denoted $\theta^\star$.
Parameter $\theta^\star$ may be chosen in a data-driven fashion to optimize
the approximation of $\Psid$ by $\Psi(\slot;\theta^\star)$; see the next subsection.
In section \ref{sec:AT} we will discuss the choice of $\theta^\star$ from the
perspective of approximation theory.

\subsection{Training and Testing}
\label{ssec:T}
The data \eqref{eq:data} is used to train the model $\Psi$; that is, to determine a choice of $\theta$.
To this end we introduce an error, or relative error, measure $r: \cV' \times \cV' \to \R^+.$ Here $\cV'$
is another Banach space containing the range of $\Psi^\dagger$ and $\Psi(\slot;\theta).$ 
Typical choices for $r$ include the error
\begin{equation}
r(v_1,v_2)=\|v_1-v_2\|_{\cV'}
\end{equation}
and, for $\varepsilon \in (0,\infty)$, one of the relative errors
\begin{equation}
r(v_1,v_2)=\frac{\|v_1-v_2\|_{\cV'}}{\varepsilon+\|v_1\|_{\cV'}}, \quad {\rm or} \quad
r(v_1,v_2)=\frac{\|v_1-v_2\|_{\cV'}}{{\rm max}\{\varepsilon,\|v_1\|_{\cV'}\}}.
\end{equation}

Now let $\muN$ be the empirical measure
$$\muN=\frac{1}{N} \sum_{n=1}^N \delta_{u_n}.$$
Then the parameter $\theta^\star$ is determined from 
\begin{align*}
\theta^*={\rm argmin}_{\theta} \,\,\cR_{N}(\theta),\quad \cR_{N}(\theta):=
\E^{u\sim\muN}\left[
r\bigl(\Psi^\dagger(u),\Psi(u;\theta)\bigr)^q
\right],
\end{align*}
for some positive $q$, typically $q=1$.
Function $\cR_{N}(\theta)$ is known as the empirical risk; also of interest is the expected (or population) risk
\[ \cR_{\infty}(\theta):=\E^{u\sim\mu}\,\left[
r\bigl(\Psi^\dagger(u),\Psi(u;\theta)\bigr)^q
\right].
\]
Note that $\cR_{\infty}(\theta)$ requires knowledge of data in the form of the entire probability measure
$\mu.$

Once trained, models are typically tested by evaluating the error 
\begin{align*}
{\sf error}:=\E^{u\sim\mu'}\left[
r\bigl(\Psi^\dagger(u),\Psi(u;\theta)\bigr)^q
\right].
\end{align*}
Here $\mu'$ is defined on the support of $\mu.$
For computational purposes the measure  $\mu'$ is often chosen 
as another empirical approximation of $\mu$,
independently of $\muN;$ other empirical measures may also be used. For theoretical analyses $\mu'$ may be chosen equal to $\mu$, but other choices may also be of interest in determining the robustness of the learned model; see, for example, \cite{benitez2023outofdistributional}.

\subsection{Finding Latent Structure}
\label{ssec:FLS}

\begin{figure}
\centering
\includegraphics[width=0.6\textwidth]{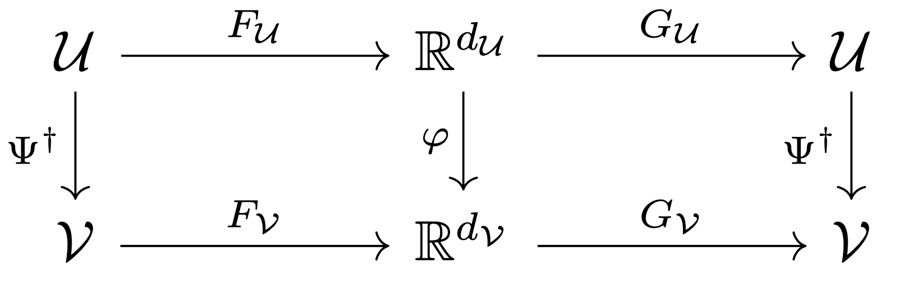}
\caption{Latent Structure in Maps Between Banach Spaces $\cU$ and $\cV$}
\label{f:3}
\end{figure}

Behind many neural operators is the extraction of latent finite dimensional structure, as illustrated
in Figure \ref{f:3}. Here we have two encoder/decoder pairs on $\cU$ and $\cV$, namely
$$G_{\cU} \circ F_{\cU} \approx I_{\cU}, \quad G_{\cV} \circ F_{\cV} \approx I_{\cV}$$
where $I_{\cU}, I_{\cV}$ are the identity maps on $\cU$ and $\cV$ respectively. Then $\varphi$ is
chosen so that
$$G_{\cV} \circ \varphi \circ F_{\cU} \approx \Psid.$$
The map $F_{\cU}$ extracts a finite dimensional latent space from the input Banach space while the map
$G_{\cV}$ returns from a second finite dimensional latent space to the output Banach space.
These encoder-decoder pairs can be learned, reducing the operator approximation to a finite
dimensional problem.

\subsection{Example (Fluid Flow in a Porous Medium)}
\label{ssec:ExD}
\begin{figure}[h]
    \begin{subfigure}{.45\textwidth}
        \includegraphics[width=\textwidth]{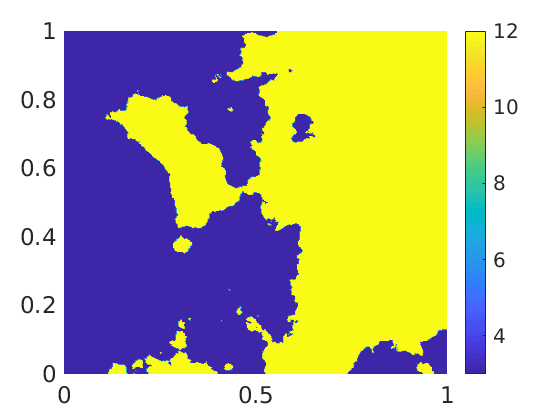}
        \caption{Input $a$}
        \label{f:4a}
    \end{subfigure} 
    ~
    \begin{subfigure}{.45\textwidth}
        \includegraphics[width=\textwidth]{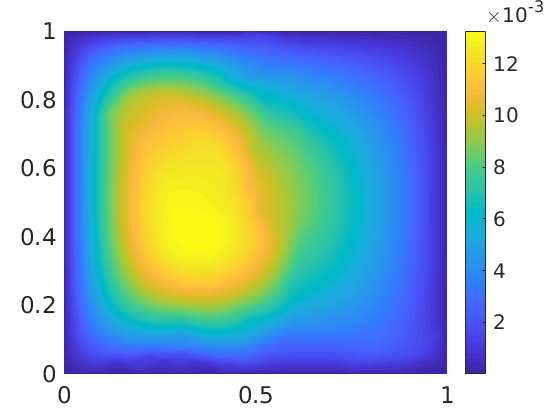}
        \caption{Output $v$}
        \label{f:4b}
    \end{subfigure}     
\end{figure}

We consider the problem of finding the piezometric head $v$ from permeability $a$
in a porous medium assumed to be governed by the Darcy Law in domain $\dD \subset \R^2.$ 
This results in the need to solve the PDE
\begin{align}
\label{eq:darcy}
\left\{
\begin{aligned}
-\nabla \cdot (a\nabla v)  & = f, \quad z \in \dD\\
v & = 0, \quad z \in \partial \dD.
\end{aligned}
\right.
\end{align} 
Here we consider $f \in H^{-1}(\dD)$ to be given and fixed. The operator of interest
\footnote{We use the notation $a$ for input functions here, because it is a commonly adopted notation in applications to porous medium flow.}
$\Psi^\dagger: a  \mapsto v$
then maps from a subset of the Banach space $L^\infty(\dD)$ into $H^1_0(\dD).$ An example of a typical
input-output pair is shown in Figures \ref{f:4a}, \ref{f:4b}. Because the equation requires strictly
positive $a$, in order  to be well-defined mathematically and to be physically meaningful, the probability measure $\mu$
must be chosen carefully. Furthermore, from the point of view of approximation theory, it is desirable
that the the space $\cU$ is separable; for this reason it is often chosen to be $L^2(\dD)$ and the measure
supported on functions $a$ satisfying a positive lower bound and a finite upper bound. Draws from such a measure
are in $L^\infty(\dD)$ and satisfy the necessary positivity and boundedness inequalities required for
a solution to the Darcy problem to exist \cite{evans2010partial}.

\section{Specific Supervised Learning Architectures}
\label{sec:Sp}

Having reviewed the general philosophy behind operator learning, we next aim to illustrate how this methodology is practically implemented. To this end, we review several representative proposals for neural operator architectures, below. 

\subsection{PCA-Net}

The PCA-Net architecture was proposed as an operator learning framework in \cite{bhattacharya2021model}, anticipated in \cite{hesthaven2018non, swischuk2019projection-based}. In the setting of PCA-Net, $\Psi^\dagger: \cU \to \cV$ is an operator mapping between Hilbert spaces $\cU$ and $\cV$, and inputs are drawn from a probability measure $\mu$ on $\cU$. Principal component analysis (PCA) is employed to obtain data-driven encoders and decoders, which are combined with a neural network mapping to give rise to the PCA-Net architecture.

The encoder is defined from PCA basis functions  on the input space $\cU$, computed from the covariance
under $\mu:$  the encoder $F_\cU$ is
determined by projection onto the first $\dx$ PCA basis functions $\{\phi_j\}_{j=1}^\dx$. The
encoding dimension $\dx$ represents a hyperparameter of the architecture. The resulting encoder is given by a linear mapping to the PCA coefficients,
\[
F_\cU: \cU \to \R^\dx, \quad F_\cU(u) = Lu := \{\left<\phi_j, u\right>\}_{j=1}^\dx. 
\]
The decoder $G_\cV$ on $\cV$ is similarly obtained from PCA under the the push-forward measure $\Psi^\dagger_\# \mu$. Denoting by $\{\psi_j\}_{j=1}^\dy$ the
first $\dy$  basis functions under this pushforward, 
the PCA-Net decoder is defined by an expansion in this basis, i.e. 
\[
G_\cU: \R^\dy \to \cV, \quad G_\cU(\alpha) = \sum_{j=1}^\dy \alpha_j \psi_j.
\] 
The PCA dimension $\dy$ represents another hyperparameter of the PCA-Net architecture.

Finally, the PCA encoding and decoding on $\cU$ and $\cV$ are combined with a finite-dimensional neural network $\alpha: \R^\dx \times \Theta \to \R^\dy$, $w \mapsto \alpha(w;\theta)$ where
$$\alpha(w;\theta) := (\alpha_1(w;\theta),\dots, \alpha_\dy(w;\theta)),$$
parametrized by $\theta \in \Theta$. This results in an operator $\Psi_{PCA}: \cU \to \cV$, of the form
\begin{align*}
    \Psi_{PCA}(u;\theta)(y) = \sum_{j=1}^{\dy}\alpha_j(L u;\theta) \psi_j(y),
\quad \forall u\in \cU, \qquad y\in \dD_y
\end{align*}
which defines the PCA-Net architecture. Hyperparameters include the dimensions of PCA $\dx$, $\dy$, and additional hyperparameters determining the neural network architecture of $\alpha$. 
In practice we are given samples
of input-/output-function pairs with $u_j$ sampled i.i.d. from $\mu$: 
$$\{(u_1,v_1),\dots, (u_N,v_N)\},$$ 
where $v_j:=\Psi^\dagger(u_j).$
Then the PCA basis functions are determined
from the covariance under an empirical approximation $\mu_N$ of $\mu$, and its pushforward under $\Psid.$ The same
data is then used to train neural network parameter $\theta \in \Theta$ defining $\alpha(w;\theta).$

\subsection{DeepONet}

The DeepONet architecture was first proposed as a practical operator learning framework in \cite{lu2021learning}, building on early work by Chen and Chen \cite{chen1995universal}. Similar to PCA-Net, the DeepONet architecture is also defined in terms of an encoder $F_\cU$ on the input space, a finite-dimensional neural network $\alpha$ between the latent finite-dimensional spaces, and a decoder $G_\cV$ on the output space.  To simplify notation, we only summarize this architecture for real-valued input and output functions. Extension to operators mapping between vector-valued functions is straightforward.

The encoder in the DeepONet architecture is given by a general linear map $L$,
\[
F_\cU: \cU \to \R^\dx, \quad F_\cU(u) = Lu.
\]
Popular choices for the encoding include a mapping to PCA coefficients, or could comprise pointwise observations $\{u(x_\ell)\}_{\ell=1}^\dx$ at a pre-determined set of so-called sensor points $x_\ell$. Another alternative of note are active subspaces \cite{zahm2020gradient}, which combine information about the input distribution and forward mapping through its gradient. This approach has been explored for function encoding in scientific ML in \cite{o2024derivative,luo2023efficient}.

The decoder in the DeepONet architecture is given by expansion with respect to a neural network basis. Given a neural network $\psi: \dD_y \times \Theta_\psi \to \R^\dy$, which defines a parametrized function from the domain $\dD_y$ of the output functions to $\R^\dy$, the DeepONet decoder is defined as
\[
G_\cV: \R^\dy \to \cV, \quad G_\cV(\alpha) = \sum_{j=1}^\dy \alpha_j \psi_j.
\]

The above encoder and decoders on $\cU$ and $\cV$ are combined with a finite-dimensional neural network $\alpha: \R^\dx \times \Theta_\alpha \to \R^\dy$, to define the parametrized DeepONet,
\begin{align*}
    \Psi_{DEEP}(u;\theta)(y) = \sum_{j=1}^{\dy}\alpha_j(Lu;\theta_{\alpha}) \psi_j(y;\theta_{\psi}), \quad \forall u\in \cU, \qquad y\in \dD_y.
\end{align*}
This architecture is specified by choice of the linear encoding $L$, and choice of neural network architectures for $\alpha$ and $\psi$. Following \cite{lu2021learning} the network $\alpha$ is conventionally referred to as the ``branch-net'' (often denoted $b$ or $\beta$), while $\psi$ is referred to as the ``trunk-net'' (often denoted $t$ or $\tau$). The combined parameters $\theta = \{ \theta_\alpha, \theta_\psi\}$ of these neural networks are learned from data of input- and output-functions.

\subsection{FNO}

The Fourier Neural Operator (FNO) architecture was introduced in \cite{li2021fourier,kovachki2023neural}. In contrast to the PCA-Net and DeepONet architectures above, FNO is not based on an approach that combines an encoding/decoding to a finite-dimensional latent space with a finite-dimensional neural network. Instead, neural operators such as the FNO generalize the structure of finite-dimensional neural networks to a function space setting, as summarized below. We will assume that the domain of the input and output functions $\dD_x = [0,2\pi]^d$ can be identified with the $d$-dimensional periodic torus.

FNO is an operator architecture of the form,
\begin{align*}
    \Psi_{FNO}(u;\theta) & = \cQ \circ \cL_{L}\circ \cdots \cL_{2}\circ \cL_{1}\circ \cR(u), \, \forall u\in \cU,\\
\cL_{\ell}(v)(x; \theta) &= \sigma\bigl(W_\ell v(x) + b_\ell+\cK(v)(x; \gamma_\ell)\bigr).
\end{align*}
It comprises of input and output layers $\cQ, \cR$, given by a pointwise composition with either a shallow neural network or a linear transformation, and several hidden layers $\cL_\ell$. 

Upon specification of a ``channel width'' $\dc$, the $\ell$-the hidden layer takes as input a vector-valued function $v: x \mapsto v(x) \in \R^\dc$ and outputs another vector-valued function $\cL_\ell(v): x \mapsto \cL_\ell(v)(x) \in \R^\dc$.\footnote{Channel width can change from layer to layer; we simplify the exposition by fixing it.} Each hidden layer involves an affine transformation 
\[
v(x) \mapsto w(x) := W_\ell v(x) + b_\ell+\cK(v)(x; \gamma_\ell),
\]
and a pointwise composition with a standard activation function 
\[
w(x) \mapsto \sigma(w(x)), 
\]
where $\sigma$ could e.g. be the rectified linear unit or a smooth variant thereof.

In the affine transformation, the matrix-vector pair $(W_\ell,b_\ell)$ defines a pointwise affine transformation of the input $v(x)$, i.e. multiplication by matrix $W_\ell$ and adding a bias $b_\ell$. $\cK$ is a convolutional integral operator, parameterized by $\gamma_\ell$,
\[
\cK(v)(x;\gamma_\ell) = \int_{\dD_x} \kappa(x-y;\gamma_\ell) v(y) \, dy,
\]
with $\kappa(\slot;\gamma_\ell)$ a matrix-valued integral kernel.
The convolutional operator can be conveniently evaluated via the Fourier transform $\cF$, giving rise to a matrix-valued Fourier multiplier,
\[
\cK(v)(x;\gamma_\ell) = \cF^{-1}( \cF(\kappa(\slot;\gamma_\ell)) \cF(v) ),
\]
where the Fourier transform is computed componentwise, and given by $\cF(v)(k) = \int_{\dD_x} v(x)e^{-\mathrm{i} k\cdot x} \, dx$.
To be more specific, if we write $\kappa(x) = (\kappa_{ij}(x))_{ij=1}^\dc$ in terms of its components, and if $\hat{\kappa}_{k,ij}$ denotes the $k$-th Fourier coefficient of $\kappa_{ij}(x)$, then the $i$-th component $\cK(v)_i$ of the vector-valued output function $\cK(v)$ is given by
\[
[\cK(v)_i](x;\gamma_\ell) = \frac{1}{(2\pi)^d}\sum_{k\in \Z^d} \left(\sum_{j=1}^\dc \hat{\kappa}_{k,ij} \cF(v_j)(k) \right) e^{\mathrm{i}k\cdot x}.
\]
Here, the inner sum represents the action of $\hat{\kappa} = \cF(\kappa)$ on $\cF(v)$, and the outer sum is the inverse Fourier transform $\cF^{-1}$. The Fourier coefficients $\hat{\kappa}_{k,ij}$ represent the tunable parameters of the convolutional operator. In a practical implementation, a Fourier cut-off $\kmax$ is introduced and the sum over $k$ is restricted to Fourier wavenumbers $|k|_{\ell^\infty}\le \kmax$, with $|\slot|_{\ell^\infty}$ the $\ell^\infty$-norm, resulting in a finite number of tunable parameters $\gamma_\ell = \{\hat{\kappa}_{k,ij}: |k|_{\ell^\infty}\le \kmax, \, i,j=1,\dots, \dc\}$.

To summarize, the FNO architecture is defined by 
\begin{enumerate}
\item an input layer $\cR$, given by pointwise composition of the input function with a shallow neural network or a linear transformation, 
\item hidden layers $\cL_1,\dots, \cL_L$ involving, for each $\ell=1,\dots, L$, matrix $W_\ell$, bias $b_\ell$ and convolutional operator $\cK(\slot;\gamma_\ell)$ with parameters $\gamma_\ell$ identified with the corresponding Fourier multipliers $\hat{\kappa}_{k,ij}$,
\item an output layer $\cQ$, given by pointwise composition with a shallow neural network or a linear transformation.
\end{enumerate}
The composition of these layers defines a parametrized operator $u \mapsto \Psi_{FNO}(u;\theta)$, where $\theta$ collects parameters from (1), (2) and (3). The parameters contained in $\theta$ are to be trained from data. The hyperparameters of FNO include the channel width $\dc$, the Fourier cut-off $\kmax$, the depth $L$ and additional hyperparameters specifying the input and output layers $\cR$ and $\cQ$. 

In theory, the FNO is formulated directly on function space and does not involve a reduction to a finite-dimensional latent space. In a practical implementation, it is usually discretized by identifying the input and output functions with their point-values on an equidistant grid. In this case, the discrete Fourier transform can be conveniently evaluated using the fast Fourier transform algorithm (FFT), and straightforward implementation in popular deep learning libraries is possible.

\subsection{Random Features Method} 

The operator learning architectures above are usually trained from data using stochastic
gradient descent. Whilst this shows great empirical success, the inability to analyze 
the optimization algorithms used by practitioners
makes it difficult to make definitive statements about the networks that are trained in practice.
The authors of \cite{nelsen2021random} have proposed a randomized alternative, by extending 
the random features methodology \cite{rahimi2007random} to a function space setting; this methodology has 
the advantage of being trainable through solution of a quadratic optimization problem.

The random feature model (RFM) requires specification of a parametrized operator $\psi: \cU \times \Gamma \to \cV$ with parameter set $\Gamma$, and a probability measure $\nu$ on the parameters $\Gamma$. Each draw $\gamma \sim \nu$ specifies a random feature $\psi(\slot;\gamma): \cU \to \cV$, i.e. a random operator. Given iid samples $\gamma_1,\dots, \gamma_M \sim \nu$, the RFM operator is then defined as
\[
\Psi_{RFM}(u; \theta)(y)
=
\sum_{j=1}^{M}\theta_{j}\psi(u;\ttheta_{j})(y)
\quad \forall u\in \cU, \, y\in \dD_y; \quad \gamma_j\,\,{\rm i.i.d.}\,.
\]
Here $\theta_1,\dots, \theta_M$ are scalar parameters. In contrast to the methodologies outlined above, the random feature model keeps the randomly drawn parameters $\gamma_1,\dots, \gamma_M$ fixed, and only optimizes over the coefficient vector $\theta = (\theta_1,\dots, \theta_M)$. With conventional loss functions, the resulting optimization over $\theta$ is convex, allowing for efficient and accurate optimization and a unique minimizer to be determined.

A suitable choice of random features is likely problem-dependent. Among others, DeepONet and FNO with randomly initialized weights are possible options. In the original work \cite{nelsen2021random}, the authors employ Fourier space random features (RF) for their numerical experiments, resembling a single-layer FNO. These Fourier space RF are specified by $\psi(u;\ttheta)=\sigma\bigl(\cF^{-1}(\chi \cF \ttheta \cF u)\bigr)$, where $\cF$ denotes Fourier transform, $\sigma$ an activation function, and $\chi$ is a Fourier space reshuffle, and $\gamma$ is drawn from a Gaussian random field.

\subsection{Discussion}
\label{ssec:disc}

The architectures above can be roughly divided into two categories, depending on how the underlying ideas from deep learning are leveraged to define a parametrized class of mappings on function space. 

\paragraph{Encoder-Decoder Network Structure} 
The first approach, which we refer to as encoder-decoder-net and which includes the PCA-Net and DeepONet architectures, involves three steps: first, the input function is encoded by a finite-dimensional vector; second, an ordinary neural network, such as a fully connected or convolutional neural network, is employed to map the encoded input to a finite-dimensional output; third, a decoder maps the finite-dimensional output to an output function in the infinite-dimensional function space. 

This approach is very natural from a numerical analysis point of view, sharing the basic structure of many numerical schemes, such as finite element methods (FEM), finite volume methods (FVM), and finite difference methods (FDM), as illustrated in Table \ref{tab:1}. From this point of view, encoder-decoder-nets mainly differ from standard numerical schemes by replacing the hand-crafted algorithm and choice of numerical discretization by a data-driven algorithm encoded in the weights and biases of a neural network, and the possibility for a data-driven encoding and reconstruction. While appealing, such structure yields approximations within a fixed, finite-dimensional, linear subspace of $\cV$. In particular, each output function from the approximate operator belongs to this linear subspace independently of the input function. Therefore these methods fall within the category of linear approximation, while methods for which outputs lie on a nonlinear manifold in $\cV$ lead to what is known as \emph{nonlinear approximation}.  The benefits of nonlinear approximation are well understood in the context of functions \cite{devore1998nonlinear}, however, for the case of operators, results are still sparse but benefits for some specific cases have been observed \cite{lee2020model, cohen2022nonlinear, lanthaler2023nonlinear, kramer2024learning}. The FNO \cite{li2021fourier} and random features \cite{nelsen2021random} are concrete examples of operator learning methodologies for which the outputs lie on a nonlinear manifold in $\cV$.

\begin{table}[h]
\begin{tabular}{c|c|c|c}
Method & Encoding & Finite-dim. Mapping & Reconstruction
\\
\hline
FEM & Galerkin projection & Numerical scheme & Finite element basis
\\
FVM & Cell averages & Numerical scheme & Piecewise polynomial
\\
FD & Point values & Numerical scheme & Interpolation 
\\
PCA-Net & PCA projection & Neural network & PCA basis
\\
DeepONet & Linear encoder & Neural network & Neural network basis
\end{tabular}
\caption{Numerical schemes vs. Encoder-Decoder-Net.}
\label{tab:1}
\end{table}

\paragraph{Neural Operators Generalizing Neural Network Structure}
A second approach to defining a parametrized class of operators on function space, distinct from encoder-decoder-nets, is illustrated by FNO. Following this approach, the structure of neural networks, which consist of an alternating composition of affine and nonlinear layers, is retained and generalized to function space. Nonlinearity is introduced via composition with a standard activation function, such as rectified linear unit or smooth variants thereof. The affine layers are obtained by integrating the input function against an integral kernel; this introduces nonlocality which is clearly needed if the
architecture is to benefit from universal approximation.

\paragraph{Optimization and Randomization}
The random features method \cite{nelsen2021random} can in principle be combined with any operator learning architecture. The random features approach opens up a less explored direction of combining optimization with randomization in operator learning. In contrast to optimization of all parameters (by gradient descent) within a neural operator approach, the RFM allows for in-depth analysis resulting in error and convergence guarantees that take into account the finite number of samples, the finite number of parameters and the optimization \cite{lanthalernelsen2023error}. One interesting, and largely unresolved question is how to design efficient random features for the operator learning setting. 

The RFM is closely related to kernel methods which have a long pedigree in machine learning. In this context, we mention a related kernel-based approach proposed in \cite{batlle2024kernel}, which employs kernel methods for operator learning within the encoder-decoder-net paradigm. This approach has shown to be competitive on several benchmark operator learning problems, and has been analyzed in \cite{batlle2024kernel}.

\paragraph{Other Approaches}

This review is mostly focused on methods that fall into one of the neural network-based approaches above, but it should be emphasized that other approaches are being actively pursued with success. Without aiming to present an exhaustive list, we mention nonlinear reduced-order modeling \cite{qian2020lift,qian2021reduced,swischuk2019projection-based,ling2016reynolds,lee2020model}, approaches based on the theory of Koopman operators \cite{yeung2019learning,peherstorfer2016data,li2017extended,morton2018deep}, work aiming to augment and speed up numerical solvers \cite{stanziola2021helmholtz}, or work on data-driven closure modeling \cite{huang2020learning,xu2021learning,liu2021a,wang2017physics,wu2018physics}, to name just a few examples. For a broader overview of other approaches to machine learning for PDEs, we refer to the recent review \cite{brunton2023machine}. While most ``operator learning'' is focused on operators mapping between functions with spatial or spatio-temporal dependence and often arising in connection with PDEs, we note that problems involving time-series represent another important avenue of machine learning research, which can also be viewed from, and may benefit from, the continuous viewpoint \cite{lanthaler2023neural}.

\section{Universal Approximation}
\label{sec:AT}

The goal of the methodologies summarized in the last section is to approximate operators mapping between infinite-dimensional Banach spaces of functions. The first theoretical question to be addressed is whether these methods can achieve this task, even in principle? The goal of this line of research is to identify classes of operators for which operator learning methodologies possess a universal approximation property, i.e. the ability to approximate a wide class of operators to any given accuracy in the absence of any constraints on the model size, the number of data samples and without any constraints on the optimization. 

Universal approximation theorems are well-known for finite dimensional neural networks
mapping between Euclidean spaces \cite{hornik1989multilayer,cybenko1989approximation},
providing a theoretical underpinning for their use in diverse applications. These results show that neural networks with non-polynomial activation can approximate very general classes of continuous (and even measurable) functions to any degree of accuracy. Universal approximation theorems for operator learning architectures provide similar guarantees in the infinite-dimensional context.

\subsection{Encoder-Decoder-Nets}
As pointed out in the last section, a popular type of architecture follows the encoder-decoder-net paradigm. Examples include PCA-Net and DeepONet. The theoretical basis for operator learning broadly, and within this paradigm more specifically, was laid out in a paper by Chen and Chen \cite{chen1995universal} in 1995, only a few years after the above cited results on the universality of neural networks. In that work, the authors introduce a generalization of neural networks, called operator networks, and prove that the proposed architecture possesses a universal property: it is shown that (shallow) operator networks can 
approximate, to arbitrary accuracy, continuous operators mapping between spaces of continuous functions. This architecture and analysis forms the basis of DeepONet, where the shallow neural networks of the original architecture of \cite{chen1995universal} are replaced by their deep counterparts. We present first a general, abstract version of an encoder-decoder-net and give a criterion on the spaces $\cU$, $\cV$ for which such architectures satisfy universal approximation. We then summarize specific results for DeepONet and PCA-Net architectures. 

We call an encoder-decoder-net a mapping $\Psi_{ED} : \cU \to \cV$ which has the form
\[\Psi_{ED} = F_{\cU} \circ \alpha \circ G_{\cV}\]
where $F_{\cU} : \cU \to \R^{d_{\cU}}$, $G_{\cV} : \R^{d_{\cV}} \to \cV$ are bounded, linear maps and $\alpha : \R^{d_{\cU}} \to \R^{d_{\cV}}$ is a continuous function. The following theorem \cite[Lemma 22]{kovachki2023neural} asserts that encoder-decoder-nets satisfy universal approximation over a large class of spaces $\cU$ and $\cV$.
\begin{theorem}
\label{thm:encoder-decoder}
Suppose that $\cU$, $\cV$ are separable Banach spaces with the approximation property. Let $\Psi^\dagger : \cU \to \cV$ be a continuous operator. Fix a compact set $K \subset \cU$. Then for any $\epsilon > 0$,
there exist positive integers $d_{\cU},d_{\cV}$, bounded linear maps $F_{\cU} : \cU \to \R^{d_{\cU}}$, $G_{\cV} : \R^{d_{\cV}} \to \cV$, and a function $\alpha \in C(\R^{d_{\cU}};\R^{d_{\cV}})$, such that
\[\sup_{u \in K} \| \Psi^\dagger (u) - (F_{\cU} \circ \alpha \circ G_{\cV})(u) \|_{\cV} \leq \epsilon.\]
\end{theorem}
A Banach space is said to have the \emph{approximation property} if, over any compact set, the identity map can be resolved as the limit of finite rank operators \cite{lindenstrauss2013classical}. Although it is a fundamental property useful in many areas in approximation theory, it is not satisfied by all separable Banach spaces \cite{enflo1973acounterexample}. However, many of the Banach spaces used in PDE theory and numerical analysis such as Lebesgue spaces, Sobolev spaces, Besov spaces, and spaces of continuously differentiable functions all posses the approximation property \cite[Lemma 26]{kovachki2023neural}. The above therefore covers a large range of scenarios in which encoder-decoder-nets can be used. We now give examples of encoder-decoder-nets where we fix the exact functional form of $F_{\cU}$ and $G_{\cV}$ and show that universal approximation continues to hold. 

\subsubsection{Operator Network and DeepONet}

The specific form of operator networks as introduced and analysed by Chen and Chen in \cite{chen1995universal} focuses on scalar-valued input and output functions. We will state the main result of \cite{chen1995universal} in this setting for notational simplicity. Extension to vector-valued functions is straight-forward. 
In simplified form, Chen and Chen \cite[cf. Theorem 5]{chen1995universal} obtain the following result:
\begin{theorem}
\label{thm:onet}
Suppose that $\sigma \in C(\R)$ is a non-polynomial activation function. Let $\dOmega \subset \R^d$ be a compact domain with Lipschitz boundary. Let $\Psi^\dagger: C(\dOmega) \to C(\dOmega)$ be a continuous operator. Fix a compact set $K\subset C(\dOmega)$.  Then for any $\varepsilon > 0$, there are positive integers $\dx,\dy,N$, sensor points $x_1,\dots, x_\dx \in \dOmega$, and coefficients $c^k_i$, $\xi^k_{ij}$, $b_i$, $\omega_k$, $\zeta_k$ with $i=1,\dots, N$, $j=1,\dots, \dx$, $k=1,\dots, \dy$, such that 
\begin{align}
\label{eq:onet}
\sup_{u\in K} 
\sup_{x\in \dOmega}
\left\vert
\Psi^\dagger(u)(x) 
- 
\sum_{k=1}^\dy \sum_{i=1}^N c_i^k \sigma\left( \sum_{j=1}^\dx \xi^k_{ij} u(x_j) + b_i\right)
\sigma(\omega_k x + \zeta_k)
\right\vert \le \varepsilon.
\end{align}
\end{theorem}

Here, we can identify the linear encoder $Lu = (u(x_1),\dots, u(x_\dx))$, the shallow branch-net $\alpha$, with components
\[
\alpha_k(Lu) = \sum_{i=1}^N c_i^k \sigma\left( \sum_{j=1}^\dx \xi^k_{ij} u(x_j) + b_i\right),
\]
and trunk-net $\psi$, with components
\[
\psi_k(y) = \sigma(\omega_k x + \zeta_k).
\]
With these definitions, \eqref{eq:onet} can be written in the equivalent form,
\[
\sup_{u\in K} \left\Vert
\Psi^\dagger(u)
- 
\sum_{k=1}^\dy \alpha_k(Lu) \psi_k
\right\Vert_{C(\dOmega)} \le \varepsilon.
\]

\begin{remark}
Theorem \ref{thm:onet} holds in much greater generality. In particular, it is not necessary to consider operator mapping input functions to output functions on the same domain $\dOmega$. In fact, the same result can be obtained for operators $\Psi^\dagger: C(V) \to C(\dOmega)$, where the input ``functions'' $u\in C(V)$ can have domain a compact subset $V$ of a general, potentially infinite-dimensional, Banach space.
\end{remark}

Theorem \ref{thm:onet} provides the motivation and theoretical underpinning for DeepONet, extended to deep branch- and trunk-nets in \cite{lu2021learning}. These results demonstrate the universality of DeepONet for a very wide range of operators, with approximation error measured in the supremum-norm over a compact set of input functions.

\paragraph{Related Work}

Several extensions and variants of DeepONets have been proposed after the initial work by Lu \emph{et al.} \cite{lu2021learning}, including extensions of the universal approximation analysis. We provide a short overview of relevant works that include a theoretical component below.

\paragraph{Input Functions Drawn From a Probability Measure} In \cite{lanthaler2022error}, Theorem \ref{thm:onet} has been generalized to input functions drawn from a general input measure $\mu$, including the case of unbounded support, such as a Gaussian measure. The error is correspondingly measured in the Bochner $L^2(\mu)$-norm (cp. discussion of PCA-Net universality below), and it is demonstrated that DeepONet can approximate general Borel measurable operators in such a setting.

\paragraph{Alternative Encoders}
There is work addressing the discretization-invariance of the encoding in DeepONet, resulting in architectures that allow for encoding of the input function at arbitrary sensor locations include Bel(Basis enhanced learning)-Net \cite{zhang2023belnet} and VIDON (Variable-input deep operator networks) \cite{prasthofer2022variable}.

The authors of \cite{HUA202321} introduce Basis Operator Network, a variant of DeepONet, where encoding is achieved by projection onto a neural network basis. Universal approximation results are obtained, including encoding error estimates for this approach. 

In \cite{jin2022mionet}, the authors address the issue of multiple input functions, and propose MIO-Net (Multiple Input/Output Net), based on tensor-product representations. The authors prove a universal approximation property for the resulting architecture, and demonstrate its viability in numerical experiments.

\paragraph{DeepONets on Abstract Hilbert Spaces} DeepONets mapping between abstract Hilbert spaces have been considered in \cite{castro2023kolmogorov,castro2022calder}, including a discussion of their universality in that context.

\subsubsection{PCA-Net} At a theoretical level, PCA-Net shares several similarities with DeepONet and much of the analysis can be carried out along similar lines. In addition to proposing the PCA-Net architecture and demonstrating its viability on numerical test problems including the Darcy flow and viscous Burgers equations, the authors of \cite{bhattacharya2021model} also prove that PCA-Net is universal for operators mapping between infinite-dimensional Hilbert spaces, with approximation error measured in the Bochner $L^2(\mu)$-norm with respect to the input measure $\mu$. This initial analysis
was developed and sharpened considerably in \cite{lanthaler2023operator}; as an example
of this we quote \cite[Proposition 31]{lanthaler2023operator}.

\begin{theorem}
[PCA-Net universality]
Let $\cU$ and $\cV$ be separable Hilbert spaces and let $\mu \in \cP(\cU)$ be a probability measure on $\cU$. Let $\Psi^\dagger: \cU \to \cV$ be a $\mu$-measurable mapping. Assume the following moment conditions,
\[
\E_{u\sim \mu}[\Vert u \Vert^2_{\cU}], \; 
\E_{u\sim \mu}[\Vert \Psi^\dagger(u) \Vert^2_{\cV}] < \infty.
\]
Then for any $\varepsilon>0$, there are dimensions $\dx$, $\dy$, a requisite amount of data $N$, a neural network $\psi$ depending on this data, such that the PCA-Net, $\Psi = G_\cV \circ \psi \circ F_\cU$, satisfies
\[
\E_{\{u_j\} \sim \mu^{\otimes N}} \left[
\E_{u\sim \mu}\left[
\Vert \Psi^\dagger(u) - \Psi(u;\{u_j\}_{j=1}^N) \Vert_{\cV}^2
\right]
\right] < \varepsilon,
\]
where the outer expectation is with respect to the iid data samples $u_1,\dots, u_N \sim \mu$, which determine the empirical PCA encoder and reconstruction.
\end{theorem}

\subsection{Neural Operators}
The Fourier neural operator (FNO) is a specific instance of a more general notion of neural operators \cite{kovachki2023neural}. The general structure of such neural operators is identical to that of the FNO, i.e. a composition
\begin{align}
\label{eq:NO}
\begin{aligned}
\Psi_{NO}(u;\theta) &= \cQ \circ \cL_L \circ \dots \circ \cL_2 \circ \cL_1 \circ \cR(u), \quad \forall \, u \in \cU, \\
\cL_\ell(v)(x;\theta) &= \sigma\left( W_\ell v(x) + b_\ell + \cK(v)(x;\gamma_\ell) \right),
\end{aligned}
\end{align}
except that the convolutional operator in each layer is replaced by a more general integral operator, 
\[
\cK(v)(x;\gamma) = \int_{\dD} \kappa(x,y; \gamma) v(y) \, dy.
\]
Here, the integral kernel $\kappa(x,y;\gamma)$ is a matrix-valued function of $x$ and $y$, parametrized by $\gamma$. Additional nonlinear dependency on the input function is possible yielding, for example, $\kappa = \kappa(x,y,u(x),u(y);\gamma)$; this structure is present in transformer models \cite{vaswani2017attention}. Different concrete implementations of such neural operators mostly differ in their choice of the parametrized integral kernel. For example, \cite{li2021fourier} uses Fourier basis, \cite{tripura2023wavelet} uses wavelet basis, and \cite{bonev2023spherical} uses spherical harmonics. Other approaches restrict the support of $\kappa$ \cite{li2020neural} or assume it decays quickly away from its diagonal \cite{li2020multipole, lam2023learning}. For a more thorough review of this methodology, we refer to \cite{kovachki2023neural}.

The universality of the FNO has first been established in \cite{kovachki2021universal}, using ideas from Fourier analysis and, in particular, building on the density of Fourier series to show that FNO can approximate a wide variety of operators. Given the great variety of possible alternative neural operator architectures, which differ from the FNO essentially only in their choice of the parametrized kernel, a proof of universality that does not explicitly rely on Fourier analysis, and which applies to a wide range of choice for the integral kernel is desirable. This has been accomplished in \cite{lanthaler2023nonlocal}, where the authors remove from the FNO all non-essential components,  from the perspective of universal approximation, yielding a minimal architecture termed the ``averaging neural operator'' (ANO).

\subsubsection{Averaging Neural Operator}
Up to non-essential details, the ANO introduced in \cite{lanthaler2023nonlocal} is a composition of nonlinear layers of the form,
\[
\cL(v;\gamma_\ell)(x) = \sigma\left( W_\ell v(x) + b_\ell(x) + V_\ell \int_{\dD} v(y) \, dy \right), \quad (\ell=1,\dots, L),
\]
where $W_\ell, V_\ell \in \R^{\dc\times \dc}$ are matrices, and $b_\ell(x)$ is a bias function. In the present work, to parametrize the bias functions, we consider bias of the form $b_\ell(x) = A_\ell x + c_\ell$ for matrix $A_\ell \in \R^{\dc \times d}$ and bias vector $c_\ell \in \R^\dc$. With this choice, the nonlinear layer $v \mapsto \cL(v)$ takes the form,
\[
\cL(v;\gamma_\ell)(x) = \sigma\left( W_\ell v(x) + A_\ell x + c_\ell + V_\ell \int_{\dD} v(y) \, dy \right).
\]
The parameter $\gamma_\ell = \{W_\ell, V_\ell, A_\ell, c_\ell\}$ collects the tunable parameters of the $\ell$-th layer. To define an operator $\Psi: \cU(D;\R^{d_i}) \to \cV(D; \R^{d_o})$, we combine these nonlinear layers with linear input and output layers $\cR: u(x) \mapsto R u(x)$ and $\cQ: v(x) \mapsto Qv(x)$, obtained by multiplication with matrices $R \in \R^{\dc \times d_i}$ and $Q\in \R^{d_o \times \dc}$, respectively. The resulting ANO is an operator of the form,
\[
\Psi(u;\theta) = \cQ \circ \cL_L\circ \dots \circ \cL_1 \circ \cR(u),
\]
with $\theta$ collecting the tunable parameters in each hidden layer, and the input and output layers. 

The ANO can be though of as a special case of FNO, where the convolutional integral kernel is constant, leading to the last term in each hidden layer being an integral or ``average'' of the input function. Similarly, the ANO is a special case of many other parametrizations of the integral kernel in neural operator architectures. Despite its simplicity, the ANO can nevertheless be shown to have a universal approximation property. We here cite a special case for operator mapping between continuous functions, and refer to \cite{lanthaler2023nonlocal} for more general results:
\begin{theorem}
\label{thm:ano-univ}
Suppose that $\sigma \in C(\R)$ is a non-polynomial activation function. Let $\dOmega \subset \R^d$ be a compact domain with Lipschitz boundary. Let $\Psi^\dagger: C(\dOmega) \to C(\dOmega)$ be a continuous operator. Fix a compact set $K\subset C(\dOmega)$. Then for any $\varepsilon> 0$, there exists an ANO $\Psi: C(\dOmega) \to C(\dOmega)$ such that 
\[
\sup_{u\in K} \Vert \Psi^\dagger(u) - \Psi(u) \Vert_{C(\dOmega)} < \varepsilon.
\]
\end{theorem}

As an immediate consequence, we have the following corollary which implies universality of neural operators for a wide range of choices:
\begin{corollary}
Consider any neural operator architecture of the form \eqref{eq:NO} with parametrized integral kernel $\kappa(x,y;\gamma)$. If for any channel dimension $\dc$ and matrix $V\in \R^{\dc\times \dc}$, there exists a parameter $\gamma_V$ such that $\kappa(x,y;\gamma_V) \equiv V$, then the neural operator architecture is universal in the sense of Theorem \ref{thm:ano-univ}.
\end{corollary}

The last corollary applies in particular to the FNO. Universality of the FNO was first established in \cite{kovachki2021universal}, there restricting attention to a periodic setting but allowing for operators mapping between general Sobolev spaces. The idea of the averaging neural operator can be found in \cite{lanthaler2023nonlocal}, where it was used to prove universality for a wide range of neural operator architectures, and for a class of operators mapping between general Sobolev spaces, or spaces of continuously differentiable functions.

\paragraph{Intuition}
We would like to provide further intuition for the universality of ANO. A simple special case of the ANO is the two-layer ANO obtained as follows. We consider neural operators which can be written as a composition of two shallow neural networks $\phi: \R^{d_i}\times \dOmega \to \R^{\dc}$, $\psi: \R^\dc \times \dOmega \to \R^{d_o}$ and an additional integral (average):
\[
\left\{
\begin{gathered}
\alpha(u) = \int_{\dOmega} \phi(u(y),y) \, dy, \\
\Psi(u)(x) = \psi\left(\alpha(u), x\right),
\end{gathered}
\right.
\]
where
\[
\phi(w,x) = C_1 \sigma(A_1 w + B_1x + d_1),
\]
and
\[
\psi(z,x) = C_2 \sigma(A_2 z + B_2 x + d_2).
\]
Note that the above composition, i.e.
\[
\Psi(u)(x) = \psi\left(\int_{\dOmega} \phi(u(y),y) \, dy, x\right),
\]
indeed defines a special case of ANO that can be written as a composition of a trivial input layer, two hidden layers and an output layer:
\[
\begin{aligned}
\cL_1(u)(x) &= \sigma\left( W_1 u(x) + b_1(x) \right), && b_1(x) = B_1 x + d_1, W_1 = A_1, \\
\cL_2(v)(x) &= \sigma\left( b_2(x) + V_2 \int v(y) \, dy \right), && b_2(x) = B_2 x + d_2, V_2 = A_2 C_1 \\
\cQ(v)(x) &= Q v(x), && Q = C_2.
\end{aligned}
\]

\begin{figure}
 
\tikzset{
pattern size/.store in=\mcSize, 
pattern size = 5pt,
pattern thickness/.store in=\mcThickness, 
pattern thickness = 0.3pt,
pattern radius/.store in=\mcRadius, 
pattern radius = 1pt}
\makeatletter
\pgfutil@ifundefined{pgf@pattern@name@_fqzr9yqis}{
\pgfdeclarepatternformonly[\mcThickness,\mcSize]{_fqzr9yqis}
{\pgfqpoint{0pt}{0pt}}
{\pgfpoint{\mcSize+\mcThickness}{\mcSize+\mcThickness}}
{\pgfpoint{\mcSize}{\mcSize}}
{
\pgfsetcolor{\tikz@pattern@color}
\pgfsetlinewidth{\mcThickness}
\pgfpathmoveto{\pgfqpoint{0pt}{0pt}}
\pgfpathlineto{\pgfpoint{\mcSize+\mcThickness}{\mcSize+\mcThickness}}
\pgfusepath{stroke}
}}
\makeatother

 
\tikzset{
pattern size/.store in=\mcSize, 
pattern size = 5pt,
pattern thickness/.store in=\mcThickness, 
pattern thickness = 0.3pt,
pattern radius/.store in=\mcRadius, 
pattern radius = 1pt}
\makeatletter
\pgfutil@ifundefined{pgf@pattern@name@_mcm93m4h1}{
\pgfdeclarepatternformonly[\mcThickness,\mcSize]{_mcm93m4h1}
{\pgfqpoint{0pt}{0pt}}
{\pgfpoint{\mcSize}{\mcSize}}
{\pgfpoint{\mcSize}{\mcSize}}
{
\pgfsetcolor{\tikz@pattern@color}
\pgfsetlinewidth{\mcThickness}
\pgfpathmoveto{\pgfqpoint{0pt}{\mcSize}}
\pgfpathlineto{\pgfpoint{\mcSize+\mcThickness}{-\mcThickness}}
\pgfpathmoveto{\pgfqpoint{0pt}{0pt}}
\pgfpathlineto{\pgfpoint{\mcSize+\mcThickness}{\mcSize+\mcThickness}}
\pgfusepath{stroke}
}}
\makeatother

 
\tikzset{
pattern size/.store in=\mcSize, 
pattern size = 5pt,
pattern thickness/.store in=\mcThickness, 
pattern thickness = 0.3pt,
pattern radius/.store in=\mcRadius, 
pattern radius = 1pt}
\makeatletter
\pgfutil@ifundefined{pgf@pattern@name@_6waqht26d}{
\pgfdeclarepatternformonly[\mcThickness,\mcSize]{_6waqht26d}
{\pgfqpoint{0pt}{0pt}}
{\pgfpoint{\mcSize+\mcThickness}{\mcSize+\mcThickness}}
{\pgfpoint{\mcSize}{\mcSize}}
{
\pgfsetcolor{\tikz@pattern@color}
\pgfsetlinewidth{\mcThickness}
\pgfpathmoveto{\pgfqpoint{0pt}{0pt}}
\pgfpathlineto{\pgfpoint{\mcSize+\mcThickness}{\mcSize+\mcThickness}}
\pgfusepath{stroke}
}}
\makeatother
\tikzset{every picture/.style={line width=0.75pt}} 

\begin{tikzpicture}[x=0.75pt,y=0.75pt,yscale=0.4,xscale=1]

\draw  [pattern=_fqzr9yqis,pattern size=6pt,pattern thickness=0.75pt,pattern radius=0pt, pattern color={rgb, 255:red, 0; green, 0; blue, 0}] (90,20) -- (130,20) -- (130,260) -- (90,260) -- cycle ;
\draw  [pattern=_mcm93m4h1,pattern size=6pt,pattern thickness=0.75pt,pattern radius=0pt, pattern color={rgb, 255:red, 0; green, 0; blue, 0}] (230,100) -- (270,100) -- (270,180) -- (230,180) -- cycle ;
\draw  [pattern=_6waqht26d,pattern size=6pt,pattern thickness=0.75pt,pattern radius=0pt, pattern color={rgb, 255:red, 0; green, 0; blue, 0}] (370,20) -- (410,20) -- (410,260) -- (370,260) -- cycle ;
\draw   (140,40) -- (220,100) -- (220,180) -- (140,240) -- cycle ;
\draw   (360,240) -- (280,180) -- (280,100) -- (360,40) -- cycle ;

\draw (82,0) node [anchor=north west][inner sep=0.75pt]    {$u( x) \in \mathcal{U}$};
\draw (170,160) node [anchor=north west][inner sep=0.75pt]    {\LARGE $\phi$};
\draw (218,50) node [anchor=north west][inner sep=0.75pt]    {$\alpha(u) \in \R^{\dc}$};
\draw (82,0) node [anchor=north west][inner sep=0.75pt]    {$u( x) \in \mathcal{U}$};
\draw (315,160) node [anchor=north west][inner sep=0.75pt]    {\LARGE $\psi$};
\draw (350,0) node [anchor=north west][inner sep=0.75pt]    {$\Psi ( u)( x) \in \mathcal{V}$};

\end{tikzpicture}
\caption{Special case of an averaging neural operator, illustrated as a nonlinear encoder-decoder architecture; with encoder $u\mapsto \alpha = \int_{\dOmega} \phi(u(y),y) \, dy$, and decoder $\alpha \mapsto \psi(\alpha,\slot)$.}
\label{fig:ano}
\end{figure}
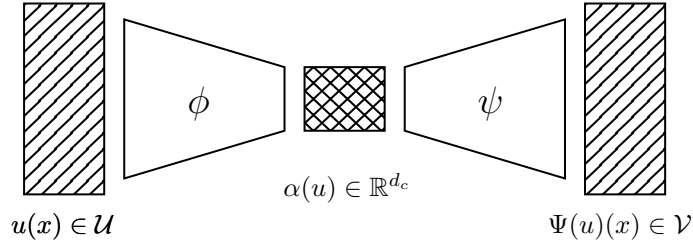

As depicted in Figure \ref{fig:ano}, such shallow ANO can be interpreted as a composition of an nonlinear encoder $\alpha: \cU \to \R^\dc$, $u \mapsto \alpha(u)$ defined via spatial averaging of $\phi$, and mapping the input function to a finite-dimensional latent space, and a nonlinear decoder $\psi: \R^\dc \to \cV$, $\alpha \mapsto \psi(\alpha,\slot)$. This interpretation opens up a path for analysis, based on which universality can be established for the ANO, and any neural operator architecture that contains such ANO as a special case, such as the Fourier neural operator.

\section{Quantitative Error and Complexity Estimates}
\label{sec:Q}

The theoretical contributions outlined in the previous section are mostly focused on methodological advances and a discussion of the universality of the resulting architectures. Universality of neural operator architectures, i.e. the ability to approximate a wide class of operators, is arguably a necessary condition for their success. But since universality is inherently qualitative, it cannot explain the efficiency of these methods in practice. Improving our understanding of the efficiency of neural operators in practice requires a quantitative theory of operator learning, providing explicit error and complexity estimates: given a desired accuracy $\varepsilon >0$, what is the model size or the number of samples that is required to achieve such accuracy? 

\subsection{Linear Operators} \label{ssec:lino}

Learning a linear operator can be formulated as solving a linear inverse problem with a non-compact forward operator \cite{mollenhauer2022learning, de2023convergence}. We take this point of view and broadly describe the results from \cite{de2023convergence}. We consider the problem of learning $\Psi^\dagger = L^\dagger$, a linear operator, when $\cU = \cV$ is a separable Hilbert space. Studying the linear problem enables
a thorough analysis of the complexity of operator learning and hence sheds light on the problem
in the more general setting. The work proceeds by assuming that $L^\dagger$ can be diagonalized in a known Schauder basis of $\cU$ denoted $\{\varphi_j\}_{j=1}^\infty$. Given input data $\{u_n\}_{n=1}^N \overset{\text{i.i.d.}}{\sim} \mu$, the noisy observations are assumed to be of the form
$$v_n = L^\dagger u_n + \gamma \xi_n$$ 
where $\gamma > 0$ and the sequence $\{\xi_n\}_{n=1}^N \overset{\text{i.i.d.}}{\sim} {\mathcal{N}(0,I_{\cU})}$ comes from a Gaussian white noise process that is independent of the input data $\{u_n\}$. In what follows here, we assume that $\mu = \mathcal{N}(0, \Gamma)$, the measure on the input data, is Gaussian with a strictly positive covariance $\Gamma$ 
and that this covariance is also diagonalizable by $\{\varphi_j\}$; note, however, that \cite{de2023convergence} treats the more general case without assuming simultaneous 
diagonalizability of $L^\dagger$ and $\Gamma$.

Note that $\{l^\dagger_j, \varphi_j\}$ uniquely determines $L^\dagger$. Thus
the problem as formulated here can be stated as learning the eigenvalue sequence 
$\{l^\dagger_j\}_{j=1}^\infty$ of $L^\dagger$ from the noisy observations
\[v_{jn} = \langle \varphi_j, u_n \rangle_{\cU} l^\dagger_j + \xi_{jn}, \quad j \in \N, \quad n = 1,\dots,N\]
where $\{\xi_{jn}\} \overset{\text{i.i.d.}}{\sim} \mathcal{N}(0,\gamma^2)$. In this problem statement, the noise is crucial for obtaining meaningful estimates on the amount of data needed for learning. Indeed, without noise, the eigenvalues can simply be recovered as 
$$l^\dagger_j = \frac{v_{j1}}{\langle \varphi_j, u_1 \rangle_{\cU}}$$ 
for all $j \in \N$ from a single data point $u_1$ since the basis $\{\varphi_j\}$ is assumed to be known. While, in practice, the observations might not be noisy, the noise process can be used to model round-off or discretization errors which occur in computation.

Assuming a Gaussian prior on the sequence $\{l^\dagger_j\} \sim \otimes_{j=1}^\infty \mathcal{N}(0, \sigma_j^2)$, one may obtain a Bayesian estimator of $L^\dagger$, given the data $\big ( \{v_{jn}\}, \{u_n\} \big )$. The Bayesian posterior is characterized as an infinite Gaussian product for the sequence of eigenvalues. We take as an estimator the mean of this Gaussian which is given as
\[l_j = \frac{\gamma^{-2} \sigma_j^2 \sum_{n=1}^N v_{jn} \langle \varphi_j, u_n \rangle_{\cU} }{1 + \gamma^{-2} \sigma_j^2 \sum_{n=1}^N \big | \langle \varphi_j, u_n \rangle_{\cU} \big |^2}.\]
Then our estimator is the operator $\Psi$, diagonalized in basis $\{\varphi_j\}$ with eigenvalues
$\{l_j\}.$ To quantify the smoothness of $L^\dagger$, we assume that $\{l^\dagger_j\}$ lives in an appropriately weighted $\ell^2$ space, in particular,
\[\sum_{j=1}^\infty j^{2 s} |l^\dagger_j |^2 < \infty\]
for some $s \in \R$. Then the following theorem holds \cite[Theorem 1.3]{de2023convergence}.

\begin{theorem}
    \label{thm:linear}
    Suppose that for some $\alpha > 1/2$ and $p \in \R$, we have $\langle \varphi_j, \Gamma \varphi_j \rangle_{\cU} \asymp j^{-2 \alpha}$ and $\sigma_j^2 \asymp j^{-2p}$. Let $\alpha' \in [0, \alpha + 1/2)$ and assume that $\min \{\alpha, \alpha'\} + \min \{p - 1/2, s\} > 0$. Then, as $N \to \infty$, we have
    \[\mathbb{E} \sum_{j=1}^\infty j^{-2 \alpha'} |l_j - l^\dagger_j|^2 \lesssim N^{- \frac{\alpha' + \min \{p - 1/2, s\} }{\alpha + p}}\]
    where the expectation is taken over the product measure defining the input data and noise.
\end{theorem}
Theorem~\ref{thm:linear} quantifies the amount of data needed, on average, for the estimator $\{l_j\}$ to achieve $\epsilon$-error in approximating $\{l^\dagger_j\}$ measured in a squared weighted $\ell^2$ norm.
In particular, we have
\[N \sim \epsilon^{- \frac{\alpha + p}{\alpha' + \min \{p - 1/2,s\}}}.\]
The exact dependence of this rate on the parameters defining the smoothness of the truth, the input, the prior, and the error metric elucidates the optimal design choices for the estimator and sheds light on which pieces are most important for the learning process. We refer to \cite{de2023convergence} for an in-depth discussion.

Within machine learning and functional data analysis, many works have focused on learning integral kernel operators \cite{rosasco2010onlearning,abernethy2009anew,crambes2013asymptotics,wang2022functional,jin2022minimax}. The operator leaning problem can then be reduced to approximation of the kernel function and is typically studied in a Reproducing Kernel Hilbert Space setting. In numerical PDEs, some recent works have studied the problem of learning the Green's function arising from some elliptic, parabolic and hyperbolic problems \cite{boulle2022data,boulle2022learning,boulle2023learning,wang2023operator}.

\subsection{Holomorphic Operators}
\label{ssec:hol}

Going beyond the linear case, holomorphic operators represent a very general class of operators for which \emph{efficient} quantitative error and complexity estimates can be established. We mention the influential work by Cohen, DeVore and Schwab \cite{cohen2010convergence,cohen2011analytic}, as well as further developments \cite{chkifa2015breaking,chkifa2013sparse,schwab2019deep,opschoor2022exponential,sz2023,herrmann2022neural,adcock2022sparse,adcock2022efficient,adcock2022near
,marcati2023exponential,adcock2023a,adcock2023b}. A detailed review, from 2015, can be found in \cite{cohen2015approximation}. We will only describe a few main ideas. We mention in passing also the works \cite{kutyniok2022theoretical,lei2022solving}, which study the neural network approximation of parametric operators in a related setting.

Holomorphic operators have mostly been studied in a parametrized setting, where the input functions can be identified with the coefficients in a suitable basis (frame) expansion. A prototypical example is the elliptic Darcy flow equation \eqref{eq:darcy}, where the coefficient field $a = a(\slot;\bm{y})$ is parametrized by a sequence $\bm{y}= (y_1,y_2,\dots) \in [-1,1]^\N$, e.g. in the form of a linear expansion,
\[
a(x;\bm{y}) = \bar{a}(x) + \sum_{j=1}^\infty y_j \varphi_j(x), \quad x \in \dOmega,
\]
where $\bar{a} \in \cU$ and $\varphi_1,\varphi_2,\dots \in \cU$ are fixed. The operator $\Psi^\dagger: a\mapsto u$ can then (loosely) be identified with the parametrized mapping,
\[
\mathsf{F}: [-1,1]^\N \to \cV, \quad \mathsf{F}(\bm{y}) := \Psi^\dagger(a(\slot;\bm{y})).
\]
In the above prototypical setting, and assuming that the sequence $\bm{b}$ with coefficients $b_j = \Vert \varphi_j \Vert_{\cV}$ decays sufficiently fast, it can be shown \cite{cohen2015approximation} that $\mathsf{F}$ possesses a holomorphic extension to a subset of the infinite product $\C^\N = \prod_{j=1}^\infty \C = \C \times \C \times \dots $ of the complex plane $\C$. More precisely, there exists a holomorphic extension
\begin{align}
\label{eq:holo}
\mathsf{F}: \prod_{j=1}^\infty \cE_{\rho_j} \to \cV_{\C},
\quad \bm{z} \mapsto \mathsf{F}(\bm{z}).
\end{align}
Here, for $\rho_j>1$, the set $\cE_{\rho_j} = \set{\frac12\left(z+z^{-1}\right) }{ z \in \C, \; 1\le |z|\le \rho_j} \subset \C$ denotes the Bernstein ellipse with focal points $\pm 1$ and major and minor semi-axes lengths $\frac12 (\rho_j\pm \rho_j^{-1})$, and $\cV_\C$ is the complexification of the Banach space $\cV$. 

In general, given a non-negative sequence $\bm{b}\in [0,\infty)^\N$ and $\varepsilon > 0$, a parametrized operator $\mathsf{F}: [-1,1]^\N \to \cV$ is called \emph{$(\bm{b},\varepsilon)$-holomorphic}, if it possesses a holomorphic extension \eqref{eq:holo} for any $\bm{\rho}=(\rho_1,\rho_2,\dots) \in [1,\infty)^\N$, s.t. 
\begin{align}
\sum_{j=1}^\infty 
\left(
\frac{\rho_j + \rho_j^{-1}}{2} - 1
\right) b_j \le \varepsilon.
\end{align}
As reviewed in \cite[Chapter 4]{adcock2022sparse}, a number of parametric differential equations of practical interest give rise to $(\bm{b},\varepsilon)$-holomorphic operators. 

The approximation theory of this class of operators is well-developed \cite{cohen2015approximation,adcock2022sparse}. The underlying reason why efficient approximation of such operators is possible is that holomorphic operators possess convergent expansions in multi-variate polynomial bases, where each polynomial basis element only depends on a finite number of the components of the complex input $\bm{z} = (z_1,z_2,\dots)$. 

A standard setting considers $\bm{b} \in \ell^p(\N)$, for some $0<p<1$. For example, if $b_j \sim j^{-s}$ decays algebraically, then $\bm{b} \in \ell^p(\N)$ for $s > p^{-1} > 1$. Assuming that $\bm{b}\in \ell^p(\N)$, it can be shown (e.g. \cite[Corollary 3.11]{cohen2015approximation}) that the best $n$-term polynomial approximation to $\mathsf{F}$ converges at rate $n^{1-1/p}$ in a sup-norm setting, and rate $n^{1/2-1/p}$ in a Bochner $L^2(\mu)$-setting, with specific input measure $\mu$ on $[-1,1]^\N$. Importantly, these convergence rates are polynomial in the number of degrees of freedom $n$, even in this infinite-dimensional parametric setting. When restricting to a finite-dimensional input space with $d$ input components, i.e. considering only inputs of the form $\bm{z} = (z_1,\dots, z_d, 0, 0, \dots)$, this fact implies that convergence rates independent of the dimension $d$ can be obtained, and thus such approximation of $(\bm{b},\varepsilon)$-holomorphic operators can provably overcome the \emph{curse of dimensionality} \cite{cohen2015approximation}.

The above mentioned results in the parametrized setting can also be used to prove efficient approximation of holomorphic operators by operator learning frameworks in a non-parametric setting \cite{lanthaler2022error,schwab2019deep,opschoor2022exponential,sz2023,herrmann2022neural}. For example, \cite{herrmann2022neural} consider the DeepONet approximation of holomorphic operators with general Riesz frame encoders and decoders, demonstrating algebraic error and complexity estimates; Under suitable conditions, the authors prove that ReLU deep operator networks (DeepONet) approximating holomorphic operators can achieve convergence rates arbitrarily close to $n^{1-s}$ in a worst-error setting (supremum norm) and at rate $n^{1/2-s}$ in a Bochner $L^2(\mu)$-norm setting. Here $n$ denotes the number of tunable parameters of the considered DeepONet, and the parameter $s$ determines the decay of the coefficients in the frame expansion of the considered input functions. Under the (loose) identification $s \sim 1/p$, these rates for DeepONet recover the rates discussed above. These results show that there exist operator surrogates which essentially achieve the approximation rates afforded by best $n$-term approximation schemes mentioned above. 

The complementary question of the sample complexity of operator learning for holomorphic operators has been studied in \cite{bachmayr2017kolmogorov,adcock2023a,adcock2023b}. Building on the theory of $N$-widths, the authors of \cite{adcock2023a,adcock2023b} show that on the class $(\bm{b},\varepsilon)$-holomorphic operators and in a Bochner $L^2$-setting, data-driven methods relying on $N$ samples cannot achieve convergence rates better than $N^{1/2-1/p}$. In addition, it is shown that the optimal rate can be achieved up to logarithmic terms. We refer to  \cite{adcock2023a,adcock2023b} for further details. 

To summarize: holomorphic operators represent a class of operators of practical interest for which approximation theory by neural operator learning frameworks can be developed. The approximation theory of this class of operators is well-developed, especially in the parametrized setting. In a parametrized setting, these operators allow for efficient approximation by multi-variate (sparse) polynomials. This fact can be leveraged to show that efficient approximation by neural network-based methods is possible, and such results can be extended beyond a parametric setting, e.g. via frame expansions. Optimal approximation rates, and methods achieving these optimal rates, up to logarithmic terms, are known under specific assumptions.

\subsection{General (Lipschitz) Operators}

The last two sections provide an overview of theoretical results on the approximation of linear and holomorphic operators. While these classes of operators include several operators of practical interest and allow for the development of general approximation theory, not all operators of relevance are holomorphic (or indeed linear). Examples of non-holomorphic operators include the solution operator associated with nonlinear hyperbolic conservation laws such as the compressible Euler equations. Solutions of such equations can develop shocks in finite time, and it can be shown that the associated solution operators themselves are not holomorphic. It is therefore of interest to extend the approximation theory of operator learning frameworks beyond the restrictive class of holomorphic operators.

A general and natural class of nonlinear operators are general Lipschitz continuous operators, the approximation theory of which has been considered from an operator learning perspective e.g. in \cite{bhattacharya2021model,liu2024deep,franco2023approximation,galimberti2022designing,schwab2023deep}. We present a brief outline of the general approach and mention relevant work on model complexity estimates in the following subsections \ref{sec:error-decomposition}--\ref{sec:cod}. Relevant results on the data complexity of operator learning in this setting are summarized in subsection \ref{ss:samplecplx}.

\subsubsection{Error Decomposition}
\label{sec:error-decomposition}

Encoder-decoder-net architectures arguably follow the basic mathematical intuition of how to approach the operator approximation problem most closely, and most theoretical work has focused on this approach. We recall that within this framework, the infinite-dimensional input and output spaces $\cU$, $\cV$ are first encoded through suitable finite-dimensional latent spaces. This involves an encoder/decoder pair $(F_\cU, G_\cU)$ on $\cU$, 
\[
F_\cU: \cU \to \R^{\dx}, \quad G_\cU: \R^{\dx} \to \cU,
\]
and another encoder/decoder pair $(F_\cV,g_\cV)$ on $\cV$,
\[
F_\cV: \cV \to \R^{\dy}, \quad G_\cV: \R^{\dy} \to \cV.
\]
We recall that the composition $G_\cU \circ F_\cU$, $G_\cV \circ F_\cV$ are interpreted as approximations to the identity on $\cU$ and $\cV$, respectively.
These encode/decoder pairs in turn imply an encoding of the underlying infinite-dimensional operator $\Psi^\dagger: \cU \to \cV$, resulting in a finite-dimensional function 
\[
\varphi: \R^\dx \to \R^\dy, \quad \varphi(\alpha) = F_\cV \circ \Psi^\dagger \circ G_\cU(\alpha),
\]
as depicted earlier, in Figure \ref{f:3}. 

While the encoder and decoder of these architectures perform dimension reduction, the neural network $\psi: \R^\dx \to \R^\dy$ at the core of encoder-decoder-net architectures is interpreted as approximating this resulting finite-dimensional function $\varphi: \R^\dx \to \R^\dy$. To summarize, an encoder-decoder-net can conceptually be interpreted as involving three steps:
\begin{enumerate}
\item Dimension reduction on the input space $\cU \approx \R^\dx$,
\item Dimension reduction on the output space $\cV \approx \R^\dy$,
\item Encoding of the operator $\Psi^\dagger$ yielding $\varphi: \R^\dx \to \R^\dy$, approximated by neural network $\psi: \R^\dx \to \R^\dy$.
\end{enumerate}
Each part of this conceptual decomposition, the encoding of $\cU \approx \R^\dx$, the decoding $\R^\dy \approx \cV$ and the approximation $\psi \approx \varphi$, represents a source of error, and the total encoder-decoder-net approximation error $\Err$ is bounded by three contributions $\Err \lesssim \Err_{\cU} + \Err_{\psi} + \Err_{\cV}$, where 
\[
\Err_{\cU} = \sup_{u} \Vert u - G_\cU \circ F_\cU(u) \Vert_{\cU},
\]
quantifies the encoding error, with supremum taken over the relevant set of input functions $u$, 
\[
\Err_{\cV} = \sup_v \Vert v - G_\cV \circ F_\cV(v) \Vert_{\cV},
\]
quantifies the decoding error, and
\begin{align}
\label{eq:err-approx}
\Err_{\psi} = \sup_{\alpha} \Vert F_\cV \circ \Psi^\dagger \circ G_\cU(\alpha) - \psi(\alpha) \Vert,
\end{align}
is the neural network approximation error.

Given this decomposition, the derivation of error and complexity estimates for encoder-decoder-net architectures thus boils down to the estimation of encoding error $\Err_{\cU}$, neural network approximation error $\Err_{\psi}$ and reconstruction error $\Err_{\cV}$, respectively.

\paragraph{Encoding and Reconstruction Errors}

Encoding and reconstruction errors are relatively well understood on classical function spaces such as Sobolev and Besov spaces, by various linear and nonlinear methods of approximation \cite{devore1993constructive}. 

For linear encoder/decoder pairs, the analysis of encoding and reconstruction errors amounts to principal component analysis (PCA) when measuring the error in the Bochner norm $L^2_\mu$, or to Kolmogorov n-widths when measuring the error in the sup-norm over a compact set. Relevant discussion of PCA in the context of operator learning is given in \cite{bhattacharya2021model} (see also \cite{lanthaler2022error} and \cite{lanthaler2023operator}).

In certain settings, such as for PDEs with discontinuous output functions, it has been shown \cite{lanthaler2023nonlinear} that relying on linear reconstruction imposes fundamental limitations on the approximation accuracy of operator methodologies, which can be overcome by methods with nonlinear reconstruction; specifically, it was shown both theoretically and experimentally in \cite{lanthaler2023nonlinear} that FNO and shift-DeepONet, a variant of DeepONet with nonlinear reconstruction, achieve higher accuracy than vanilla DeepONet for prototypical PDEs with discontinuous solutions. We also mention closely related work on the nonlinear manifold decoder architecture of \cite{seidman2022nomad}.

\paragraph{Neural Network Approximation Error}

At their core, encoder-decoder-net architectures employ a neural network to approximate the encoded version $F_\cV \circ \Psi^\dagger \circ G_\cU$ of the underlying operator $\Psi^\dagger$ (cp. \eqref{eq:err-approx}). The practical success of these frameworks thus hinges on the ability of ordinary neural networks to approximate the relevant class of high-dimensional functions in the latent-dimensional spaces, which are obtained through the encoding of such operators.
While the empirical success of neural networks in high-dimensional approximation tasks is undeniable, our theoretical understanding and the mathematical foundation underpinning this empirical success remains incomplete. 

General approximation theoretic results on the neural network approximation of functions have been obtained, and some available quantitative bounds in operator learning \cite{franco2023approximation,galimberti2022designing} build on these results to estimate the neural network approximation error $\Err_\psi$. Notably, the seminal work \cite{yarotsky2017error} of D. Yarotsky presents general error and complexity estimates for functions with Lipschitz continuous derivatives:
\begin{theorem}
\label{thm:nn-cod}
A function $f \in W^{k,\infty}([0,1]^d)$ can be approximated to uniform accuracy $\varepsilon > 0$,
\[
\sup_{x\in [0,1]^d} | f(x) - \psi(x)|\le \varepsilon,
\]
by a ReLU neural network $\psi$ with at most $O(\varepsilon^{-d/k} \log(\varepsilon^{-1}))$ tunable parameters. 
\end{theorem}

\begin{remark}
\label{rem:nn-scaling}
Note that the relevant dimension in the operator learning context is the latent dimension $d = d_\cU$. Neglecting logarithmic terms, we note that each component of the function $G: \R^\dx \to \R^\dy$ can be approximated individually by a neural network of size at most $O(\varepsilon^{-\dx/k})$, and hence, we expect that $G$ can be approximated to accuracy $\varepsilon$ by a neural network $\psi$ of size at most $O(\dy \varepsilon^{-\dx/k})$.
\end{remark}

Without aiming to provide a comprehensive overview of this very active research direction on neural network approximation theory, adjacent to operator learning theory, we mention that similar error and complexity estimates can also be obtained on more general Sobolev spaces, e.g. \cite{yarotsky2020phase,siegel2023optimal}. Lower bounds illuminating the limitations of neural networks on model classes are for example discussed in \cite{achour2022a,yarotsky2017error,bolcskei2019optimal}. Approximation rates leveraging additional structure beyond smoothness have also been considered, e.g. compositional structure is explored in \cite{mhaskar2020function,shen2019nonlinear,sh2020
}.

\paragraph{Non-standard Architectures and Hyperexpressive Activations} 

While the general research area of neural network approximation theory is too broad to adequately summarize here, we mention relevant work on hyperexpressive activations, which can formally break the curse of dimensionality observed Theorem \ref{thm:nn-cod}; it has been shown that neural networks employing non-standard activations can formally achieve arbitrary convergence on model function classes \cite{pinkus1999approximation,yarotsky2021elementary,shen2021neural
,liang2021reproducing}, when the complexity is measured in terms of number of tunable parameters. This is not true for the ReLU activation \cite{yarotsky2017error}. Another way to break the curse of dimensionality is via architectures with non-standard ``three-dimensional'' structure \cite{zhang2022neural}. 

While non-standard architectures can overcome the curse of dimensionality in the sense that the number of parameters does not grow exponentially with $d$ (or is independent of $d$), it should be pointed out that this necessarily comes at the expense of the number of bits that are required to represent each parameter in a practical implementation. Indeed, from work on quantized neural networks \cite{bolcskei2019optimal} (with arbitrary activation function), it can be inferred that the total number of bits required to store all parameters in such architectures is lower bounded by the Kolmogorov $\varepsilon$-entropy of the underlying model class; For the specific model class $W^{k,\infty}([0,1]^d)$, this entropy scales as $\varepsilon^{-d/k}$. Hence, architectures which achieve error $\varepsilon$ with a number of parameters that scales strictly slower than $\varepsilon^{-d/k}$ must do so at the expense of the precision that is required to represent each individual parameter in a practical implementation, keeping the total number of bits above the entropy limit. For related discussion, we e.g. refer to \cite[section 7]{yarotsky2020phase} or \cite[discussion on page 5]{siegel2023optimal}. Another implication of this fact is that the constructed non-standard architectures are necessarily very sensitive to minute changes in the network parameters.

\subsubsection{Upper Complexity Bounds}
\label{sec:upper}

Quantitative error estimates for operator learning based on the general approach outlined in the last subsection \ref{sec:error-decomposition} have been derived in a number of recent works \cite{liu2024deep,franco2023approximation,galimberti2022designing,bhattacharya2021model,HUA202321,MHASKAR2023194}.

\paragraph{Relevant work}
The two papers \cite{bhattacharya2021model,lanthaler2022error}, analyzing PCA-Net and DeepONet respectively, both introduce a splitting of the error into encoder, neural network approximation and reconstruction errors.
A similar error analysis is employed in \cite{HUA202321} for so-called ``basis operator network'', a variant of DeepONet. An in-depth analysis of DeepONets with various encoder/decoder pairs, including generalization error estimates, is given in \cite{liu2024deep}. Quantitative approximation error estimates for convolution neural networks applied to operator learning are derived in \cite{franco2023approximation}. General error estimates motivated by infinite-dimensional dynamical systems in stochastic analysis can be found in \cite{galimberti2022designing}. An alternative approach to operator learning with explicit algorithms for all weights is proposed in \cite{MHASKAR2023194},  including error estimates for this approach.

\paragraph{Alternative Decompositions}
Finally, we point out that while the error decomposition in encoding, neural network approximation and reconstruction errors is natural, alternative error decompositions, potentially more fine-grained, are possible. We mention the work \cite{patel2024variationally} which proposes a mimetic neural operator architecture inspired by the weak variational form of elliptic PDEs, discretized by the finite-element method; starting from this idea, the authors arrive at an architecture that can be viewed as a variant of DeepONet, including a specific mixed nonlinear and linear branch network structure and a nonlinear trunk net. In this work, an a priori error analysis is conducted resulting in a splitting of the overall approximation in numerical approximation, stability, training and quadrature errors depending on the data-generation with a numerical scheme (no access to the actual operator), the Lipschitz stability of the underlying operator, the finite number of training samples and quadrature errors to approximate integrals, respectively.

\subsubsection{Lower Complexity Bounds}
\label{sec:cod}

Operator learning frameworks are based on neural networks and provide highly nonlinear approximation \cite{devore1998nonlinear}. Despite their astonishing approximation capabilities, even highly nonlinear approximation methods have intrinsic limitations.

\paragraph{Combined Error Analysis for Encoder-Decoder-Nets}

To illustrate some of these intrinsic limitations, we first outline the combined error analysis that results from the decomposition summarized in the last section. To this end, we combine the encoding, reconstruction and neural network analysis to derive quantitative error and complexity estimates within the encoder-decoder-net paradigm.

Firstly, under reasonable assumptions on the input functions, the encoding error can often be shown to decay at an algebraic rate in the $d_\cU$, e.g.
\begin{align}
\label{eq:eu}
\Err_{\cU} \lesssim d_{\cU}^{-\alpha}.
\end{align}
For example, if we assume that the input functions are defined on a bounded domain $\dOmega \subset \R^d$ and subject to a smoothness constraint such as a uniform bound on their $k$-th derivative, then a decay rate $\alpha = k/d$ can be achieved (depending on the precise setting); For dimension reduction by principal component analysis, the exponent $\alpha$ instead relates to the decay rate of the eigenvalues of the covariance operator of the input distribution.

Under similar assumptions on the set of output functions, depending on the properties of the underlying operator $\Psi^\dagger$, the reconstruction error on the output space often also decays algebraically,
\begin{align}
\label{eq:ev}
\Err_{\cV} \lesssim d_{\cV}^{-\beta},
\end{align}
where the decay rate $\beta$ can e.g. be estimated in terms of the smoothness of the output functions under $\Psi^\dagger$, or could be related to the decay of the PCA eigenvalues of the output distribution (push-forward under $\Psi^\dagger$).

Finally, given latent dimensions $d_{\cU}$ and $d_\cV$, the size of the neural network $\psi$ that is required to approximate the encoded operator mapping $G: \R^\dx \to \R^\dy$, with NN approximation error bound,
\[
\Err_\psi \le \varepsilon,
\]
roughly scales as (cp. Remark \ref{rem:nn-scaling}),
\begin{align}
\label{eq:epsi}
\size(\psi) \sim \dy \varepsilon^{-\dx/k},
\end{align}
when the only information on the underlying operator is captured by its degree of smoothness $k$ ($k=1$ corresponding to Lipschitz continuity). Note that this is the scaling consistent with Kolmogorov entropy bounds.

Given the error decomposition $\Err \lesssim \Err_{\cU} + \Err_{\psi} + \Err_{\cV}$, we require each error contribution individually to be bounded by $\varepsilon$. In view of \eqref{eq:eu} and \eqref{eq:ev}, this can be achieved provided that $\dx \sim \varepsilon^{-1/\alpha}$, $\dy \sim \varepsilon^{-1/\beta}$. Inserting such choice of $\dx$, $\dy$ in \eqref{eq:epsi}, we arrive at a neural network size of roughly the form,
\[
\size(\psi) \sim \varepsilon^{-1/\beta} \varepsilon^{-c\varepsilon^{-1/\alpha}/k}.
\]
In particular, we note the exponential dependence on $\varepsilon^{-1}$, resulting in a size estimate,
\begin{align}
\label{eq:cod}
\size(\psi)
\gtrsim
\exp\left(
\frac{c\varepsilon^{-1/\alpha}}{k}
\right).
\end{align}
As pointed out after \eqref{eq:eu}, when the set of input functions consists of functions defined on a $d$-dimensional domain with uniformly bounded $s$-th derivatives (in a suitable norm), then we expect a rate $\alpha = s/d$, in which case we obtain,
\begin{align}
\label{eq:cod1}
\size(\psi)
\gtrsim
\exp\left(
\frac{c\varepsilon^{-d/s}}{k}
\right).
\end{align}

For operator learning frameworks, this super-algebraic (even exponential) dependence of the complexity on $\varepsilon^{-1}$ has been termed the ``curse of dimensionality'' in \cite{kovachki2021universal,lanthaler2022error
} or more recently ``curse of parametric complexity'' in \cite{lanthaler2023operator}. The latter term was introduced to avoid confusion, which may arise because in these operator learning problems, there is no fixed dimension to speak of. The curse of parametric complexity can be viewed as an infinite-dimensional scaling limit of the finite-dimensional curse of dimensionality, represented by the $d_\cU$-dependency of the bound $\varepsilon^{-d_\cU/k}$, and arises from the finite-dimensional CoD by observing that the required latent dimension $\dx$ itself depends on $\varepsilon$, with scaling $\dx \sim \varepsilon^{-1/\alpha}$. We note in passing that even if $\dx \sim \log(\varepsilon^{-1})$ were to scale only logarithmically in $\varepsilon^{-1}$, the complexity bound implied by \eqref{eq:epsi} would still be super-algebraic, consistent with the main result of \cite{lanthaler2023operator}.

\paragraph{Nonlinear $n$-width Estimates}

The rather pessimistic complexity bound outlined in \eqref{eq:cod} is based on an upper bound on the operator approximation error $\Err$, and is not necessarily tight. One may therefore wonder if more careful estimates could yield complexity bounds that do not scale exponentially in $\varepsilon^{-1}$. 

In this context, we would like to highlight the early work on operator approximation by Mhaskar and Hahm \cite{mhaskar1997neural} which presents first quantitative bounds for the approximation of nonlinear functionals; most notably, this work identifies the continuous nonlinear $n$-widths of spaces of H\"older continuous functionals defined on $L^2$-spaces; it is shown that the relevant $n$-widths decay only (poly-)logarithmically in $n$, including both upper and lower bounds.

We will presently state a simplified version of the main result of \cite{mhaskar1997neural}, and refer to the original work for the general version. To this end, we recall that the continuous nonlinear $n$-width $d_\mathcal{N}(\cK;\Vert \slot\Vert_{\cX})$ \cite{devore1989optimal} of a subset $\cK \subset \cX$, with $(\cX,\Vert \slot \Vert_{\cX})$ Banach, is defined as the optimal reconstruction error,
\[
d_\mathcal{N}(\cK;\Vert \slot\Vert_{\cX})
= 
\inf_{(a,M)} \sup_{f\in \cK} \Vert f - M(a(f)) \Vert_{\cX},
\]
where the infimum is over all encoder/decoder pairs $(a,M)$, consisting of a continuous map $a: \cK \to \R^n$ and general map $M: \R^n \to \cX$.

To derive lower $n$-width bounds, we consider spaces of nonlinear Lipschitz functionals $\Psi^\dagger \in \mathfrak{F}_{d}$, where $d$ denotes the spatial dimension of the input functions. More, precisely define 
\[
\mathfrak{F}_{d}
=
\set{
\Psi^\dagger: L^2([-1,1]^d) \to \R
}{\Vert \Psi^\dagger(u) \Vert_{\Lip} \le 1
},
\]
with 
\[
\Vert \Psi^\dagger \Vert_{\Lip} 
:= 
\sup_{u\in L^2} |\Psi^\dagger(u)|
+
\sup_{u,v\in L^2} \frac{| \Psi^\dagger(u) - \Psi^\dagger(v) |}{\Vert u - v\Vert_{L^2}}.
\]
Given a smoothness parameter $s>0$, we consider approximation of $\Psi^\dagger \in \mathfrak{F}_d$, uniformly over a compact set of input functions $K^s\subset L^2([-1,1]^d)$, obtained as follows: we expand input functions $f\in L^2([-1,1]^d)$ in a Legendre expansion, $f(x) = \sum_{k\in \N^d} \hat{f}_k P_k(x)$, and consider functionals defined on the ``Sobolev'' ball,
\[
K^s := \set{f\in L^2([-1,1]^d)}{\sum_{k\in \N^d} |k|^{2s} \hat{f}_k|^2 \le 1}.
\]
We measure the approximation error between $\Psi^\dagger, \Psi: K^s \subset L^2 \to \R$ with respect to the supremum norm over $K^s$,
\[
\Vert \Psi^\dagger - \Psi \Vert_{C(K^s)} 
=
\sup_{u\in K^s} |\Psi^\dagger(u) - \Psi(u)|.
\]
It follows from the main results of \cite{mhaskar1997neural} that the continuous nonlinear $n$-widths of the set of functionals $\mathfrak{F}_d$ decay only (poly-)logarithmically, as
\[
d_n(\mathfrak{F}_{d})_{C(K^s)} 
\sim \log(n)^{-s/d}.
\]
In particular, to achieve uniform approximation accuracy $\varepsilon>0$ with a \emph{continuous} encoder/decoder pair $(a,M)$, requires at least
\begin{align}
\label{eq:nwidth}
n \gtrsim \exp\left(c\varepsilon^{-d/s}\right),
\end{align}
parameters. This last lower bound should be compared with \eqref{eq:cod1} (for $k=1$); the lower $n$-width bound \eqref{eq:nwidth} would imply \eqref{eq:cod1} under the assumption that the architecture of $\psi = \psi(\slot;\theta)$ was fixed and assuming that the weight assignment $\Psi^\dagger \mapsto \theta_{\Psi^\dagger}$ from the functional $\Psi^\dagger$ and the optimal tuning of neural network parameters $\theta_{\Psi^\dagger}$ was continuous. The latter assumption may not be satisfied if parameters are optimized using gradient descent. 

\paragraph{Curse of Parametric Complexity}

Given the result outlined in the previous paragraph, one may wonder if the pessimistic bound \eqref{eq:nwidth} and \eqref{eq:cod}, i.e. the ``curse of parametric complexity'', can be broken by (a) a dis-continuous weight assignment, and (b) an adaptive choice of architecture optimized for specific $\Psi^\dagger$. This question has been raised in \cite{lanthaler2023operator,lanthaler2023curse}.

It turns out that, with operator learning frameworks such as DeepONet, FNO or PCA-Net, and relying on standard neural network architectures, it is not possible to overcome the curse of parametric complexity when considering approximation on the full class of Lipschitz continuous or Fr\'echet differentiable operators. We mention the following illustrative result for DeepONet, which follows from \cite[Example 2.17]{lanthaler2023curse}:

\begin{proposition}[Curse of Parametric Complexity]
Let $\dOmega \subset \R^d$ be a domain. Let $k \in \N$ be given, and consider the compact set of input functions,
\[
K = \set{u \in C^k(\dOmega)}{\Vert u \Vert_{C^k} \le 1} \subset \cU := C(\dOmega).
\]
Fix $\alpha > 2 + \frac{k}{d}$. Then for any $r\in \N$, there exists a $r$-times Fr\'echet differentiable functional $\Psi^\dagger: \cU \to \R$ and constant $c,\bar{\varepsilon}>0$, such that approximation to accuracy $\varepsilon \le \bar{\varepsilon}$ by a DeepONet $\Psi: \cU \to \R$ with ReLU activation,
\begin{align}
\label{eq:accuracy}
\sup_{u \in K} \vert \Psi^\dagger(u) - \Psi(u) \vert \le \varepsilon,
\end{align}
with linear encoder $\cE$ and neural network $\psi$, implies complexity bound 
\begin{align}
\label{eq:cod2}
\size(\psi) \ge \exp(c \varepsilon^{-1/\alpha r}).
\end{align}
Here $c$, $\bar{\varepsilon}>0$ are constants depending only on $k$, $\alpha$ and $r$.
\end{proposition}

As mentioned above, analogous lower complexity bounds can be obtained for PCA-Net, Fourier neural operator and many other architectures, and the more general version of this lower complexity bound applies to Sobolev input functions and beyond. We refer to \cite{lanthaler2023curse} for a detailed discussion.

\begin{remark}
In \cite{lanthaler2023curse}, no attempt was made to optimize the exponent of $\varepsilon^{-1}$ in \eqref{eq:cod2}. It would be interesting to know whether the appearance of the degree of operator F\'echet differentiability, i.e. the parameter $r$, in the lower bound is merely an artifact of the proof in \cite{lanthaler2023curse}. The back-of-the-envelope calculation leading to \eqref{eq:cod} indicates that $r$ should not appear in the exponent, and that the factor $\alpha = k/d+2+\delta$ could be replaced by $k/d+\delta$.
\end{remark}

\paragraph{Breaking the Curse of Parametric Complexity with Non-Standard Architectures}
As mentioned in a previous section, there exist non-standard neural network architectures which either employ non-standard activations \cite{pinkus1999approximation,yarotsky2021elementary,shen2021neural}, or impose a non-standard ``three-dimensional'' connectivity \cite{zhang2022neural} which can overcome the curse of dimensionality in finite dimensions. In particular, encoder-decoder-nets based on such non-standard architectures can achieve neural network approximation error $\Err_{\psi}\le \varepsilon$ with a complexity (as measured by the number of tunable degrees of freedom) that grows much slower than the rough scaling we considered in \eqref{eq:epsi}. Based on such architectures, it has recently been shown \cite{schwab2023deep} that DeepONets can achieve approximation rates for general Lipschitz and H\"older continuous operators which break the curse of parametric complexity implied by \eqref{eq:cod2}. In fact, such architectures achieve algebraic expression rate bounds for general Lipschitz and H\"older continuous operators.

\subsubsection{Sample Complexity Results}
\label{ss:samplecplx}

There is a rapidly growing body of work on the approximation theory of operator learning with focus on parametric complexity. The complementary question of the sample complexity of operator learning, i.e. how many samples are needed to achieve a given approximation accuracy, has not received as much attention.
The work described in subsection \ref{ssec:lino} addresses this question in the setting of learning
linear operators, and the question is also addressed in subsection \ref{ssec:hol} for holomorphic operators. 
We now develop this subject further. Of particular note in the general Lipschitz setting of this subsection is the paper \cite{liu2024deep}, as well as related recent work in \cite{chen2023deep}, which studies the nonparametric error estimation of Lipschitz operators for general encoder-decoder-net architectures. In \cite{liu2024deep}, non-asymptotic upper bounds for the generalization error of empirical risk minimizers on suitable classes of operator networks are derived. The results are stated in a Bochner $L^2(\mu)$ setting with input functions drawn from a probability measure $\mu$, and variants are derived for general (Lipschitz) encoder/decoder pairs, for fixed basis encoder/decoder pairs, and for PCA encoder/decoder pairs. The analysis underlying the approximation error estimates in \cite{liu2024deep} is based on a combination of best-approximation error estimates (parametric complexity) which are combined with statistical learning theory to derive sample complexity bounds. 

For detailed results applicable to more general settings, we refer the reader to \cite{liu2024deep}. Here, we restrict attention to a representative result for fixed basis encoder/decoder pair \cite[Corollary 3]{liu2024deep}, obtained by projection onto a trigonometric basis.

To state this simplified result, consider a Lipschitz operator $\Psi^\dagger: \cU \to \cV$, mapping between spaces $\cU, \cV = L^2([-1,1]^d)$. We assume that there exists a constant $C>0$, such that the probability measure $\mu\in \cP(\cU)$ and its push-forward $\Psi^\dagger_\# \mu \in \cP(\cV)$ are supported on periodic, continuously differentiable functions belonging to the set,
\[
K := \set{u \in L^2([-1,1]^d)}{\text{$u$ is periodic}, \; \Vert u \Vert_{C^{k,\alpha}} \le C}.
\]
Then the squared approximation error 
\[
\Err^2 := 
\E_{\text{data}} \E_{u\sim \mu} 
\Vert 
D_\cY \circ \psi \circ E_\cX(u) - \Psi^\dagger(u)
\Vert_{L^2}^2 
\]
satisfies the upper bound,
\begin{align}
\label{eq:genest}
\Err^2
\lesssim \dy^{\frac{4+\dx}{2+\dx}} N^{-\frac{2}{2+\dx}} \log^6(n) + \dx^{-\frac{2s}{d}} + \dy^{-\frac{2s}{d}},
\end{align}
where $N$ is the number of samples and $s = k + \alpha$ is the smoothness on the input and output spaces. The neural network $\psi$ is a ReLU network of depth $L$ and width $p$, satisfying (up to logarithmic terms),
\[
Lp \sim
\dy^{\frac{\dx}{4+2\dx}}
N^{\frac{\dx}{4+2\dx}}.
\]
Comparing with \eqref{eq:eu} and \eqref{eq:ev}, we can identify the last two terms in \eqref{eq:genest} as the encoding and reconstruction errors. The first term corresponds to a combination of neural network approximation and generalization errors.

To ensure that the total error $\Err \le \varepsilon$ is below accuracy threshold $\varepsilon$, we first choose $\dx, \dy \sim  \varepsilon^{-d/s}$, consistent with our discussion in subsection \ref{sec:cod}. And according to the above estimate, we choose a number of samples of roughly the size $N \sim \varepsilon^{-(2+\dx)/2}$. Note that, once more, the additional $\varepsilon$-dependency of $\dx$ implies that 
\[
N \gtrsim \exp\left(c \varepsilon^{-d/s}\right),
\]
exhibits an exponential curse of complexity. This time, the curse is reflected by an exponential number of samples $N$ that are required to achieve accuracy $\varepsilon$, rather than the parametric complexity. In turn, this implies that the size of the product $Lp$ of the depth $L$ and width $p$ of the neural network $\psi$ satisfies the lower bound,
\[
Lp \gtrsim \exp\left( c \varepsilon^{-d/s}\right),
\]
consistent with the expected curse of parametric complexity, \eqref{eq:cod}. It is likely that the results of \cite{liu2024deep} cannot be substantially improved in the considered setting of Lipschitz operators. Extending their work to a slightly different setting, the authors of \cite{liu2024deep} also raise the question of low dimensional structure in operator learning, and derive error bounds decaying with a fast rate under suitable conditions, relying on low-dimensional latent structure of the input space.

In a related direction, the authors of \cite{kim2022bounding} provide estimates on the Rademacher complexity of FNO. Generalization error estimates are derived based on these Rademacher complexity estimates, and the theoretical insights are compared with the empirical generalization error and the proposed capacity of FNO, in numerical experiments. Out-of-distribution risk bounds for neural operators with focus on the Helmholtz equation are discussed in depth in \cite{benitez2023outofdistributional}.

We finally mention the recent work \cite{mukherjee2023size}, where a connection is made between the number of available samples $n$ and the required size of the DeepONet reconstruction dimension $\dy$. It is shown that when only noisy measurements are available, a scaling of the number of trunk basis functions $\dy \gtrsim \sqrt{n}$ is required to achieve accurate approximation.

\subsection{Structure Beyond Smoothness}

The results summarized in the previous sections indicate that, when relying on standard neural network architectures, \emph{efficient} operator learning on general spaces of Lipschitz continuous, or Fr\'echet differentiable, operators may not be possible: the class of all such operators on infinite-dimensional Banach spaces is arguably too rich, and operator learning on this class suffers from a curse of parametric complexity, requiring exponential model sizes of the form $\gtrsim \exp(c\varepsilon^{-\gamma})$.

This is in contrast to operator learning for $(\bm{b},\varepsilon)$-holomorphic operators, for which approximation to accuracy $\varepsilon$ is possible with a parametric complexity $O(\varepsilon^{-\gamma})$ scaling only algebraically in $\varepsilon^{-1}$. In this case, the curse of parametric complexity is broken by the extraordinary amount of smoothness of the underlying operators, going far beyond Lipschitz continuity or Fr\'echet differentiability.

These contrasting results rely only on the smoothness of the approximated operator: Is such smoothness the deciding factor for the practical success of operator learning methodologies? While we currently cannot provide a theoretical answer to this important question, we finally would like to mention several approximation theoretic results addressing how operator learning frameworks can break the curse of parametric complexity by leveraging structure beyond holomorphy.

\paragraph{Operator Barron Spaces}
A celebrated result in the study of shallow neural networks on finite-dimensional spaces is Barron's discovery \cite{barron1993universal} of a function space on which \emph{dimension-independent} Monte-Carlo approximation rates $O(1/\sqrt{n})$ can be obtained. In particular, the approximation, by shallow neural networks,  of functions belonging to this Barron class does not suffer from the well-known curse of dimensionality.

In the recent paper \cite{korolev2021two}, a suitable generalization of the Barron spaces is introduced, and it is shown that Monte-Carlo approximation rates $O(1/\sqrt{n})$ can be obtained even in this infinite-dimensional setting, under precisely specified conditions. Quantitative error estimates (convergence rates) for the approximation of nonlinear operators are obtained by extending earlier results \cite{bach2017breaking,ma2022barron,wojtowytsch2022representation
} from the finite-dimensional setting $f:\R^d \to \R$ to the vector-valued and infinite-dimensional case $f:\cU  \rightarrow  \cV$, where $\cU$  and $\cV$ are Banach spaces. 

The operator Barron spaces identified in \cite{korolev2021two} represent a general class of operators, distinct from the holomorphic operators discussed in a previous section, which allow for efficient approximation by a class of ``shallow neural operators''. Unfortunately, a priori, it is unclear which operators of practical interest belong to this class, leaving the connection between these theoretical results and the practically observed efficiency of neural operator somewhat tenuous. In passing, we also mention the operator reproducing kernel Hilbert spaces (RKHS) considered in the context of the random feature model in \cite{nelsen2021random}, for which similar Monte-Carlo convergence rates have been derived in \cite{lanthalernelsen2023error}.

\paragraph{Representation Formulae and Emulation of Numerical Methods}

To conclude our discussion of complexity and error bounds, we mention work focused on additional structure, separate from smoothness and the above-mentioned idea of Barron spaces, which can be leveraged by operator learning frameworks to achieve efficient approximation: these include operators with explicit representation formulae, and operators for which efficient approximation by traditional numerical schemes is possible. Such representations by classical methods can often be efficiently emulated by operator learning methodologies, resulting in error and complexity estimates that beat the curse of parametric complexity. 

The complexity estimates for DeepONets in \cite{deng2021convergence} are mostly based on explicit representation of the solution; most prominently, this is achieved via the Cole-Hopf transformation for the viscous Burgers equation. 

Results employing emulation of numerical methods to prove that operator learning frameworks such as DeepONet, FNO and PCA-Net can overcome the curse of parametric complexity for specific operators of interest can be found in \cite{kovachki2021universal,lanthaler2022error,lanthaler2023operator,marcati2023exponential}; specifically, such results have e.g. been obtained for the Darcy flow equation, the Navier-Stokes equations, reaction-diffusion equations and the inviscid Burgers equation. For the solution operators associated with these PDEs, it has been shown that operator learning frameworks can achieve approximation accuracy $\varepsilon$ with a total number of tunable degrees of freedom which either only scales algebraically in $\varepsilon$, i.e. with $\size(\psi) = O(\varepsilon^{-\gamma})$, or scales only logarithmically,  $\size(\psi) = O(|\log \varepsilon |^{\gamma})$ in certain settings \cite{marcati2023exponential}. This should be contrasted with the general curse of dimensionality \eqref{eq:cod}. It is expected that the underlying ideas apply to many other PDEs.

Results in this direction are currently only available for very specific operators, and an abstract characterization of the relevant structure that can be exploited by operator learning frameworks is not available. First steps towards a more general theory have been proposed in \cite{ryck2022generic}, where generic bounds for operator learning are derived, relating the approximation error for physics-informed neural networks (PINNs) and operator learning architectures such as DeepONets and FNOs.

\subsubsection{Discussion}

Ultimately, the overarching theme behind many of the above cited results is that neural operators, or neural networks more generally, can efficiently emulate numerical algorithms, which either result from bespoke numerical methods or are a consequence of explicit representation formulae. The total complexity of a neural network emulator, and reflected by its size, is composed of the complexity of the emulated numerical algorithm and an additional overhead cost of emulating this algorithm by a neural network (translation to neural network weights). From an approximation-theoretic point of view, it could be conjectured that, for a suitable definition of ``numerical algorithm'', neural networks can efficiently approximate \emph{all numerical algorithms}, hence implying efficient approximation of a great variety of operators, excluding only those operators for which no efficient numerical algorithms exist. Formalizing a suitable notion of numerical algorithm and proving that neural networks can efficiently emulate any such algorithm would be of interest and could provide a general way for proving algebraic expression rate bound for a general class of operators that can be approximated by a numerical method with algebraic memory and run-time complexity (i.e. any ``reasonable'' approximation method).


\section{Conclusions}
\label{sec:C}

Neural operator architectures employ neural networks to approximate nonlinear operators mapping between Banach spaces of functions. Such operators often arise from physical models which are expressed as partial differential equations (PDEs). 
Despite their empirical success in a variety of applications, our theoretical understanding of neural operators remains incomplete. This review article summarizes recent progress and the current state of our theoretical understanding of neural operators, focusing on an approximation theoretic point of view.

The starting point of the theoretical analysis is universal approximation. Very general universal approximation results are now available for many of the proposed neural operator architectures. These results demonstrate that, given a sufficient number of parameters, neural operators can approximate a very wide variety of infinite-dimensional operators, providing a theoretical underpinning for diverse applications. Such universal approximation is arguably a necessary but not sufficient condition for the success of these architectures. In particular, universal approximation is inherently qualitative and does not guarantee that approximation to a desired accuracy is feasible at a practically realistic model size.

A number of more recent works thus aim to provide quantitative bounds on the required model size and the required number of input-/output-samples to achieve a desired accuracy $\epsilon$. Most such results consider one of three settings: general Lipschitz (or Fr\'echet differentiable) operators, holomorphic operators, or specific PDE operators. While Lipschitz operators are a natural and general class to consider, it turns out that approximation to error $\epsilon$ with standard architectures requires an exponential (in $\epsilon^{-1}$) number of tunable parameters, bringing into question whether operator learning at this level of generality is possible. In contrast, the class of holomorphic operators allows for complexity bounds that scale only algebraically in $\epsilon^{-1}$, both in terms of models size as well as sample complexity. Holomorphic operators represent a general class of operators of practical interest, for which rigorous approximation theory has been developed building on convergent (generalized) polynomial expansions.

Going beyond notions of operator smoothness, it has been shown that operator learning frameworks can leverage intrinsic structure of (PDE-) operators to achieve algebraic convergence rates in theory; this intrinsic structure is distinct from holomorphy. Available results in this direction currently rely a case-by-case analysis and often leverage emulation of traditional numerical methods. The authors of the present article view the development of a general approximation theory, including a characterization of the relevant structure that can be leveraged by neural operators, as one of the great challenges of this field.

\paragraph{Acknowledgments.} The authors are grateful to Dima Burov and Edo Calvello for creating Figures \ref{f:1a} and \ref{f:1b}, \ref{f:2a}, \ref{f:2b} respectively. They are also grateful to Nick Nelsen for reading, and commenting on, a draft of the paper. NBK is grateful to the NVIDIA Corporation for full time employment. SL is grateful for support from the Swiss National Science Foundation, Postdoc.Mobility grant P500PT-206737. AMS is grateful for support from a Department of Defense Vannevar Bush Faculty Fellowship.

\bibliographystyle{abbrv}
\bibliography{references}

\end{document}